\documentclass[11pt]{article}

\usepackage[final]{acl}

\usepackage{times}
\usepackage{latexsym}
\usepackage{amsmath}

\usepackage[T1]{fontenc}

\usepackage[utf8]{inputenc}

\usepackage{microtype}
\usepackage{multirow}

\usepackage{inconsolata}

\usepackage{graphicx}
\graphicspath{ {./images/} }
\usepackage{algorithm}
\usepackage{algpseudocode}

\title{Identifying Temporal Features within Transcoders for Time Sensitive Factual Recall}

\author{Sanjay Govindan, Maurice Pagnucco, Yang Song \\
        University of New South Wales, Sydney}

\newcommand{\commontoyear}{\textit{common to the year}}
\newcommand{\commontemporal}{\textit{common temporal}}
\begin{document}
\maketitle
\begin{abstract}
Large Language Models (LLMs) suffer from temporal misalignment, often due to the contradictory nature of their training corpora. While current mitigation strategies rely on computationally expensive fine-tuning or context-heavy retrieval augmented generation (RAG), the internal mechanisms governing time sensitive recall remain under-explored. Unlike prior studies that identify temporal components such as attention heads and MLP layers, we provide the first feature level map of temporal recall by isolating individual MLP features via transcoder circuit tracing. We identify three node categories (\commontemporal{}, \commontoyear{} and \textit{chrono-semantic}) which interact to generate a temporal filter during factual recall.  By analysing Gemma 2 2B, LLaMA 3.2 1B, and Qwen3-4b, we show that these features do not follow a simple linear pipeline but represent time through a parallel and mixed syntactic-semantic interplay across layers. We additionally discover a class of higher-layer temporal components invisible to existing EAP-IG methods, establishing transcoders as a more complete lens for temporal interpretability for time sensitive factual recall. These findings present MLP components for potential targeted interventions in time-sensitive factual recall. Our code can be found at \url{https://github.com/sg-sy/timewarp-temporaltranscoder}.
\end{abstract}

\section{Introduction}
Large language models (LLMs) are trained with a chaotic sense of time \citep{Zhao2024-li,Luu2022-fk}. Their pre-training and post-training data contain a wide range of statements that are both true and untrue depending on the temporal context \citep{jiang2025time, vu-etal-2024-freshllms}. When answering the question "who is the current President of the United States of America?", the training data is most likely to contain a range of answers throughout time (e.g., George Washington, Barack Obama, Donald Trump). Given the training corpus has a direct influence on the probability of output tokens, the idea of the "current" US president can be skewed by the popularity of a president within the corpus, leading to misalignment for temporally relative statements \citep{Geva2023-le,Dhingra2022-in, cho-etal-2025-understanding, ou-etal-2025-llms, 10.5555/3692070.3692115}.

Altering the temporal alignment of models has traditionally been achieved through in-context learning and fine-tuning methods. These methods produce better temporal alignment but come with their own drawbacks. In the case of in-context learning, performance comes at the cost of enlarged context windows, and with the use of retrieval augmented generation (RAG) \citep{Lewis2020-wi}, indexation and retrieval issues \citep{Zhang2025-gq, pham-etal-2024-whos}. On the other hand, fine-tuning requires an increase in time and computational resources, whilst also producing a static model that is highly temporally aligned \citep{Zhao2024-li, li-etal-2024-revisiting}. Both approaches incur significant costs: bloated context windows, convoluted retrieval mechanisms, or catastrophic forgetting effects \citep{Zhang2025-gq,Zhao2024-li, doi:10.1073/pnas.1611835114, liu-niehues-2025-conditions}. 

To address the internal temporal misalignment of models, prior studies \citep{govindan-etal-2025-temporal,Park2025-ok} have tried to identify time relevant circuitry and align models using activation engineering techniques. Their results are mixed, demonstrating model components that can intermittently toggle temporal concern for time sensitive questions, highlighting the complex and opaque nature of time sensitive factual recall.

To provide a more complete view of temporal component identification and information access for time sensitive factual recall \citep{blackboxnlp-ws-2025-1, Luu2022-fk}, we identify neural components and circuitry that are critical to the correct recall of time sensitive facts; an extension of knowledge circuits \citep{Park2025-ok, DBLP:journals/corr/abs-2405-17969, lindsey2025biology}. We achieve this through the combination of transcoders and circuit analysis techniques, specifically transcoder circuit tracing \citep{ameisen2025circuit, circuittracer, lindsey2025biology}. We use transcoder circuit tracing to create, analyse and identify temporal sub-graphs, producing a variety of interpretable features. Our research identifies three sets of core temporal components, \commontemporal{}, \commontoyear{} and \textit{chrono-semantic} features. By examining the effects on ablating and steering critical nodes from \commontemporal{} and \commontoyear{}, we find that there are temporally continuous \textit{chrono-semantic} features that are not identified as temporal components using EAP-IG methods, but are a strong temporal reference. Our findings are a fundamental step towards disentangling factual recall and temporal representations, highlighting how time is represented as a set of continuous temporal anchors.

\begin{figure*}
  \centering
  \includegraphics[width=400px]{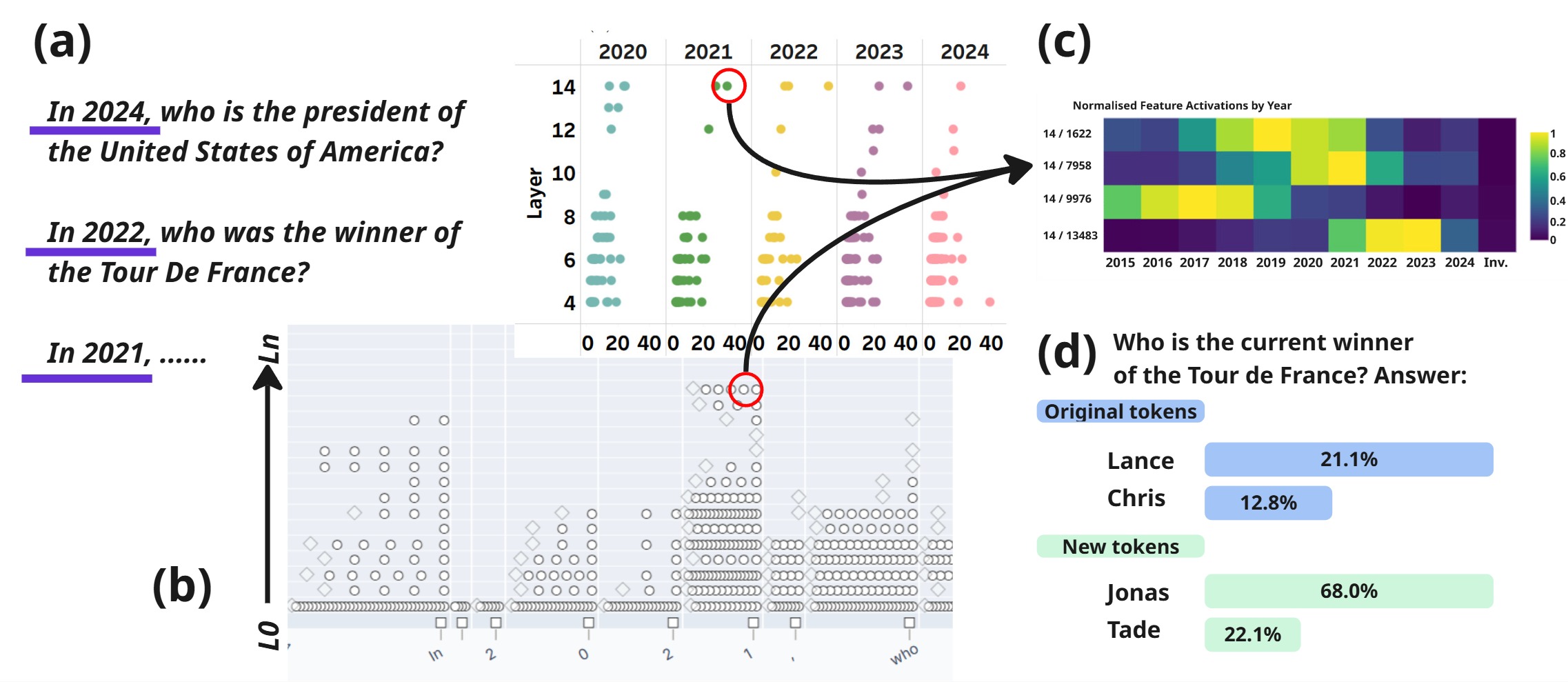}
  \caption{The overview of our experimental setup. \textbf{(a)} Using a new dataset, ChronosAlign, to generate question answer pairs over time. \textbf{(b)} illustrates the construction of transcoder circuits across a range of time sensitive question answers. This enables a refined understanding of which layers and transcoder features are selectively important for specific years  (\commontoyear{} features), or temporal recall (\commontemporal{} features). \textbf{(c)} We validate our findings by testing certain nodes against time sensitive subjects (people, places and events) and noting the average token activation for specific nodes and years. \textbf{(d)} Finally, we demonstrate the importance of these nodes on time sensitive recall through ablation and steering studies, using logit probability differences and F1 scores to measure outcomes - aligning to 2022 in this example}
  \label{fig:IntroductionOverview}
\end{figure*}

\section{Preliminaries}
In the following sections we introduce the datasets and the background of circuit analysis and transcoders, which are key tools in identifying and altering the temporal understanding of LLMs.

\subsection{Datasets}
\label{sec:datasets}

\textbf{Time sensitive - ChronosAlign} To facilitate temporal alignment across models with varying training cutoff periods, we developed ChronosAlign which provides a framework and dataset for continuous evaluation of temporal alignment as models evolve. We identify time-sensitive question-answer pairs using Wikidata and Wikipedia textual and tabular data. Our dataset contains approximately 26k questions with up to 500k question-answer pairs spanning 2010--2025. This study centres on a 2020--2024 timeframe for its core ablation and steering experiments, containing \textasciitilde{}14k unique questions, and \textasciitilde{}70k question answer pairs over the period. Whilst we focus on the 2020--2024 period for ablation and steering we utilise the 2015--2025 data to provide broader temporal context in our visualisations.

\textbf{Time-invariant}
To assess the impact of temporal feature ablation and steering on non-temporal reasoning, we incorporate Commonsense \citep{Hernandez2024-hx} and GSM8K \citep{cobbe2021gsm8k} datasets to measure effects on numerical reasoning tasks.

\subsection{Circuit Analysis}
\label{sec:CircuitAnalysis}
Circuit analysis is a mechanistic interpretability technique that aims to identify which model components causally contribute to specific outputs. The approach represents neural networks as directed acyclic graphs, enabling researchers to trace information flow and isolate the computational pathways responsible for particular behaviours. In this framework, nodes ($N$) correspond to model components such as embeddings ($I$), attention heads ($A_{l,h}$), MLPs ($M_l$), and outputs ($O$), while edges ($E$) capture the information flow between these components. Formally:
\begin{equation}
N = \{I, A_{l,h}, M_l, O\}
\end{equation}
\begin{equation}
E = \{ (n_x, n_y) \mid n_x, n_y \in N \}
\end{equation}

\noindent Prior studies \citep{Park2025-ok, marks2025sparse} have employed this representation to identify key sub-circuits ($C \subseteq (N,E)$) underlying various model capabilities.

\subsection{Transcoders}
Neurons exhibit polysemantic behaviour, encoding multiple unrelated features. This polysemy obscures the reasons behind specific neuron activations. Where circuit analysis (Section \ref{sec:CircuitAnalysis}) highlights the critical components, transcoder circuits aim to uncover the ``why'' behind the importance. The goal of a transcoder is to take in $W_{in}$, and reconstruct $W_{out}$, mimicking the original MLP component, but with a larger dimensionality. Formally, a transcoder computes:

\begin{equation}
\mathbf{f} = \sigma(W_{\text{enc}} \mathbf{x} + \mathbf{b}_{\text{enc}})
\end{equation}
\begin{equation}
\hat{\mathbf{y}} = W_{\text{dec}} \mathbf{f} + \mathbf{b}_{\text{dec}}
\end{equation}

\noindent where $\mathbf{x}$ is the input activation vector to the MLP sublayer, $W_{\text{enc}}$ is the encoder weight matrix, $\mathbf{b}_{\text{enc}}$ is the encoder bias vector, $\sigma$ is a nonlinear activation function (typically ReLU) to induce sparsity, $\mathbf{f}$ is the sparse feature activation vector, $W_{\text{dec}}$ is the decoder weight matrix, and $\mathbf{b}_{\text{dec}}$ is the decoder bias vector. The interpretability comes from the wide and sparse reconstruction of the MLP node (denoted as $f$), leading to a simplified and monosemantic interpretable layer \citep{NEURIPS2024_2b8f4db0}.

\section{Temporal transcoder feature identification}

We explore how transcoder circuit analysis can provide steerable insight into time sensitive knowledge recall circuitry. We focus on uncovering MLP components and transcoder representations that are critical to the temporal component of time sensitive recall; we are identifying what information moved, and not why it moved \cite{kamath2025tracing}. Raw transcoder circuits generate thousands of feature nodes per prompt. Identifying which nodes genuinely encode temporal information requires systematic decomposition across three phases. Specifically, we first identify critical nodes via influence scoring. We then categorise these nodes by their activation patterns across temporal contexts. Finally, we validate identified features through ablation and steering interventions.

\subsection{Temporal circuit construction} 
Circuits can be created from transcoder nodes if we consider the impact these feature nodes' have on the final output logits. The links between these nodes are represented as edges, where the weights of the edges can be calculated if we make assumptions and freeze some of the non-linear components such as attention and layer norm \citep{ameisen2025circuit}. Removing these non-linear components simplifies the source-to-output logit calculation to a set of linear equations; for any given feature and node, we can now efficiently calculate the impact of that node upon the output logits. We create the indirect influence matrix ($B$) by obtaining the adjacency matrix of the graph ($A$) and calculating the summed product of the weighted edges between two nodes (Eq. \ref{eq:TranscoderCircuits1}), which can be simplified to Eq. \ref{eq:TranscoderCircuits2}.

\begin{equation}
\label{eq:TranscoderCircuits1}
B = A + A^2 + A^3 + \cdots
\end{equation}
\begin{equation}
\label{eq:TranscoderCircuits2}
B = (I - A)^{-1} - I
\end{equation}
\begin{equation}
\label{eq:TranscoderCircuits3}
S_i = \sum_{j} p_j \cdot B_{ij}, \quad p_j = \frac{e^{z_j}}{\sum_{k} e^{z_k}}
\end{equation}

\noindent A final logit influence score (Eq. \ref{eq:TranscoderCircuits3}) is derived
by weighting the indirect influences by the model's output probabilities, where $S_i$ is the influence score for node $i$, $B_{ij}$ is the indirect influence of node $i$ on output logit $j$ from Eq. \ref{eq:TranscoderCircuits2}, and $p_j$ is the softmax probability assigned to token $j$ over logits $z$. This final score provides an efficient mechanism for pruning transcoder circuits as it isolates nodes that causally affect model outputs. This filtering constrains analysis from thousands of initial candidates to tractable subsets.


\begin{figure*}
\centering
\begin{minipage}[t]{0.48\textwidth}
\centering
\includegraphics[width=\textwidth,height=5cm,keepaspectratio]{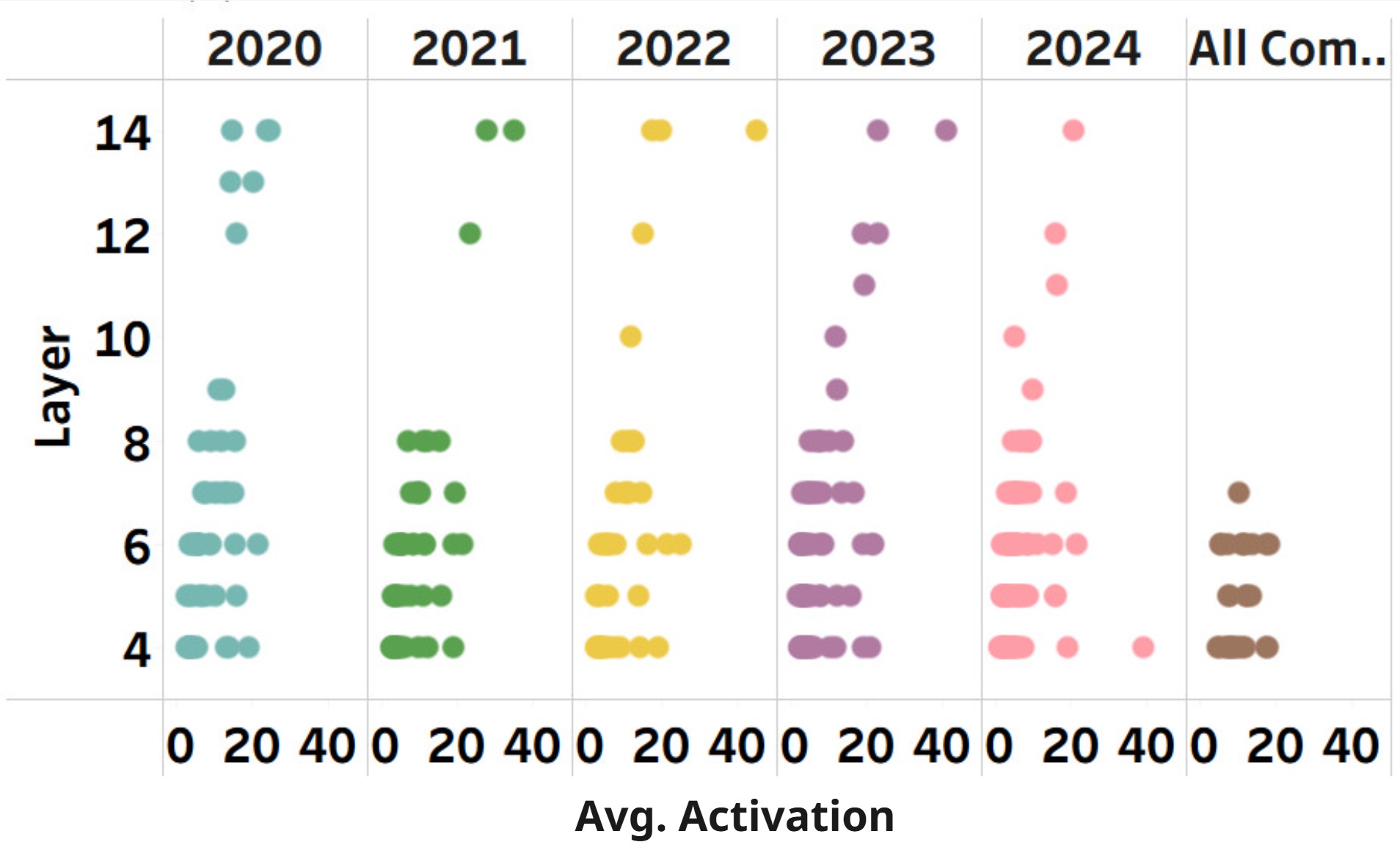}
\caption{Distribution of temporal nodes across model layers and their average activations. The \commontemporal{} (brown) features between layers 4-7, compared to \commontoyear{} features which highlight a combination of syntactic and temporal semantic features. One notable trend is the increase in nodes between layers 8 and 14 as we approach and exceed the training data cutoff period.}
\label{fig:TemporalNodes_LayersAndActivationMapping}
\end{minipage}
\hfill
\begin{minipage}[t]{0.48\textwidth}
\centering
\includegraphics[width=\textwidth,height=5cm,keepaspectratio]{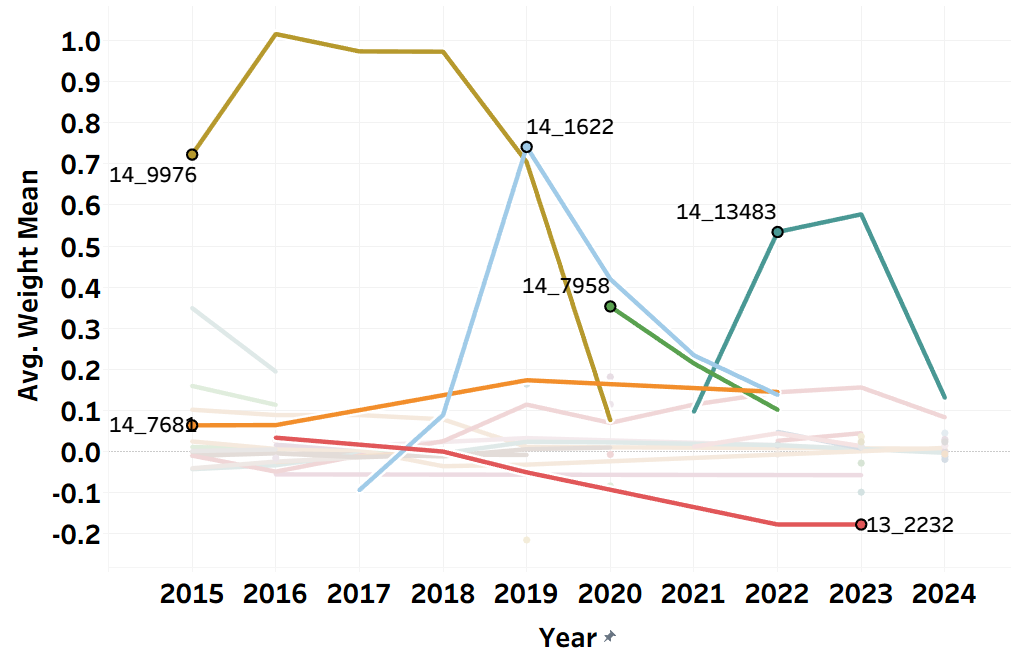}
\caption{Gemma 2, higher layer nodes. Mean edge weights (y-axis), by years (x-axis) highlighting the overlap of strong temporal anchors (L14F9976, L14F1622, L14F13483) across a 2015 to 2024 continuum. We highlight high weighted edge nodes, and contrast that against some low weighted edge nodes, demonstrating the scaled representation of time across a range of higher layer nodes.}
\label{fig:TemporalNodes_FeatureMapping}
\end{minipage}
\end{figure*}

Not all high-influence nodes relate to temporal grounding. We thus isolate specific temporal tokens and conduct an overlap analysis to partition nodes into functionally distinct categories. This reveals which nodes encode truly temporal versus coincidental features. To isolate genuinely temporal circuitry, we apply token-level filters, focusing analysis exclusively on positions corresponding to year expressions within prompts (e.g., In 2020 ...). This constraint reduces spurious node activation from semantic features orthogonal to temporal grounding. Subsequently, we employ a node overlap analysis (Eq. \ref{eq:graph_iteration}, \ref{eq:node_categorization_year}, \ref{eq:node_categorization_temporal} ) to systematically partition identified nodes into functionally distinct categories:

\begin{equation}
\label{eq:graph_iteration}
\forall G_i \in \mathcal{G}, \quad \text{where} \quad \mathcal{G} = \{G_1, G_2, \ldots, G_n\}
\end{equation}


\begin{equation}
  \label{eq:node_categorization_year}
C_y = \bigcap_{\substack{G_i \in \mathcal{G} \\ \text{year}(G_i) = y}} N_i
\end{equation}

\begin{equation}
\label{eq:node_categorization_temporal}
C_t = \bigcap_{i=1}^{n} N_i
\end{equation}


\noindent where $\mathcal{G}$ represents the collection of all generated circuit graphs from a set of prompts, $N_i$  denotes the set of nodes in graph $G_i$. This decomposition yields two node categories: (1) \commontoyear{} nodes ($C_y$ ) that reliably activate only for specific year ($y$); (2) \commontemporal{} nodes ($C_t$) that consistently activate across all temporal contexts, forming a potential backbone for temporal representation.

We employ circuit construction to systematically identify temporal features across a curated set of question-answer pairs from ChronosAlign. For each of three semantic domains (sport, politics, culture), we sample up to 20 representative high-confidence cases over 2020-2024. Final influence scores (Eq. \ref{eq:TranscoderCircuits3}) constrain the initial node set to approximately 17,000 candidates per temporal prompt, enabling tractable downstream analysis. Experiments are conducted across three LLMs, Gemma 2 2B \citep{gemma2024}, LLaMA 3.2 1B \citep{Grattafiori2024-nw} and  Qwen3-4b \citep{qwen3technicalreport}, using per layer transcoders for each of the models \citep{lieberum2024gemmascopeopensparse}, with a primary focus of Gemma 2 2B.

\subsection{Causal intervention methodology}
\label{sec:evaluation metrics}
To investigate the validity of our temporal node identification, we employ two complementary intervention strategies. \textit{Temporal ablation} suppresses \commontemporal{} features, and conversely, \textit{temporal steering} amplifies \commontoyear{} features. Both interventions follow the formulation:

\begin{equation}
\label{eq:steering}
A_{p,l} = m \cdot (a + n_{p,l})
\end{equation}

\noindent where $p$ and $l$ denote token position and layer index respectively and $n$ represents the activation at $p,l$. $m$ and $a$ are the multiplicative and additive coefficients that are used for ablation or steering.

To measure the intervention impact, we adopt a dual-metric evaluation framework to capture complementary dimensions of intervention efficacy. While logit based metrics directly reflect feature causal influence on model outputs, F1 accuracy-based metrics validate that observed shifts correspond to semantically coherent predictions. This combination is necessary because intervention-induced logit changes alone may not reflect genuine temporal recall improvements if they produce incoherent answers.

\textbf{Logit probability shifts for relative, invariant and explicit answers} Temporal feature interventions should differentially affect how strongly the model commits to relative and invariant versus explicit temporal framings. We measure four complementary quantities: (1) explicit logit diff (change in probability of explicit answers), (2) invariant logit diff (change in probability of invariant answers), (3) prior inv-vs-exp (denoted as \textit{ive}) difference (invariant minus explicit probabilities before intervention), and (4) post-intervention ive difference (invariant minus explicit probabilities after intervention). For metrics 2,3 and 4, we also construct a relative version of these metrics, denoted as \textit{rve}. These metrics isolate feature contributions to the model's temporal mode selection.

\textbf{F1 accuracy} To ensure interventions produce valid responses rather than arbitrary logit shifts, we measure answer correctness via F1 scores against invariant, relative and explicit model generated answers. Since smaller models may lack complete ground-truth knowledge \citep{Park2025-ok}, we evaluate against both explicit and relative answer distributions rather than absolute truth. This metric prevents interpreting semantically nonsensical logit shifts as successful temporal feature manipulation.

\section{Results and Discussion}

We organise our investigation around three questions:
(\S~\ref{sec:TemporalFeaturesClusterbyProcessingStage}) What features assist in time sensitive question answer recall? 
(\S~\ref{sec:ablation}) Are these features causally necessary? 
(\S~\ref{sec:steering}) Can they be steered to control outputs?
We primarily focus on Gemma 2 2B, but present findings for both LLaMA 3.2 1B and Qwen3-4b in the results and in Appendix \ref{app:ablation_Qwen3-4b} \& \ref{app:ablation_LLaMA 3.2 1B}.

\subsection{Temporal Feature Representations}
\label{sec:TemporalFeaturesClusterbyProcessingStage}

If temporal recall is compositional, we expect features to stratify by abstraction level: lower layers should encode syntactic temporal patterns (numerical year representations), while higher layers should integrate these with semantic content \citep{Geva2023-le}. However, the boundary between these stages and their potential overlap remains unclear. Our decomposition reveals three functionally distinct feature types distributed across model layers (Figure \ref{fig:TemporalNodes_LayersAndActivationMapping}):

\begin{enumerate}
    \item \textbf{Common temporal features (Layers 4--7):} Generic numerical temporal representations that activate across all time periods (e.g., ``years starting with 20,'' ``four-digit numbers representing years'')
    
    \item \textbf{Common to the year features (Layers 4+):} Year-specific numerical anchors that reliably activate for particular temporal contexts 
    (e.g., features selective for 2020 vs. 2024)
    
    \item \textbf{Chrono-semantic Nodes (Layers 4+):} For a subset of \textbf{Common to the year} features that link temporal markers to semantic content. (e.g., \texttt{L7F6200}: ``coronavirus pandemic'', \texttt{L4F14100}: ``names of politicians and celebrities'', \texttt{L14F13483}: ``terms related to stock market activity and economic indicators'')
\end{enumerate}

\textit{Parallel processing, not strict hierarchy.} Figure 
\ref{fig:TemporalNodes_LayersAndActivationMapping} reveals substantial overlap between \commontoyear{} and \commontemporal{} features in layers 4--7. Rather than sequential processing that filters on specific point in time (e.g., common year $\rightarrow$ specific year $\rightarrow$ temporal anchor), temporal information appears to flow through parallel streams that converge in mid-layers temporal representations. This suggests that year-specific and \commontemporal{} features are computed simultaneously and integrated compositionally.

\textit{Early semantic anchoring.} Contrary to the hypothesis that semantic temporal content emerges only in higher layers \citep{Geva2023-le}, we observe event-anchored features as early as layer 4 (\texttt{L4F14100}: names of contemporary public figures) and distinct temporal events by layer 7 (\texttt{L7F6200}: coronavirus pandemic language). This early emergence of high-level temporal concepts suggests that models do not cleanly separate syntactic and semantic temporal processing.

\begin{figure}[hbt!]
  \centering
  \includegraphics[width=1\linewidth]{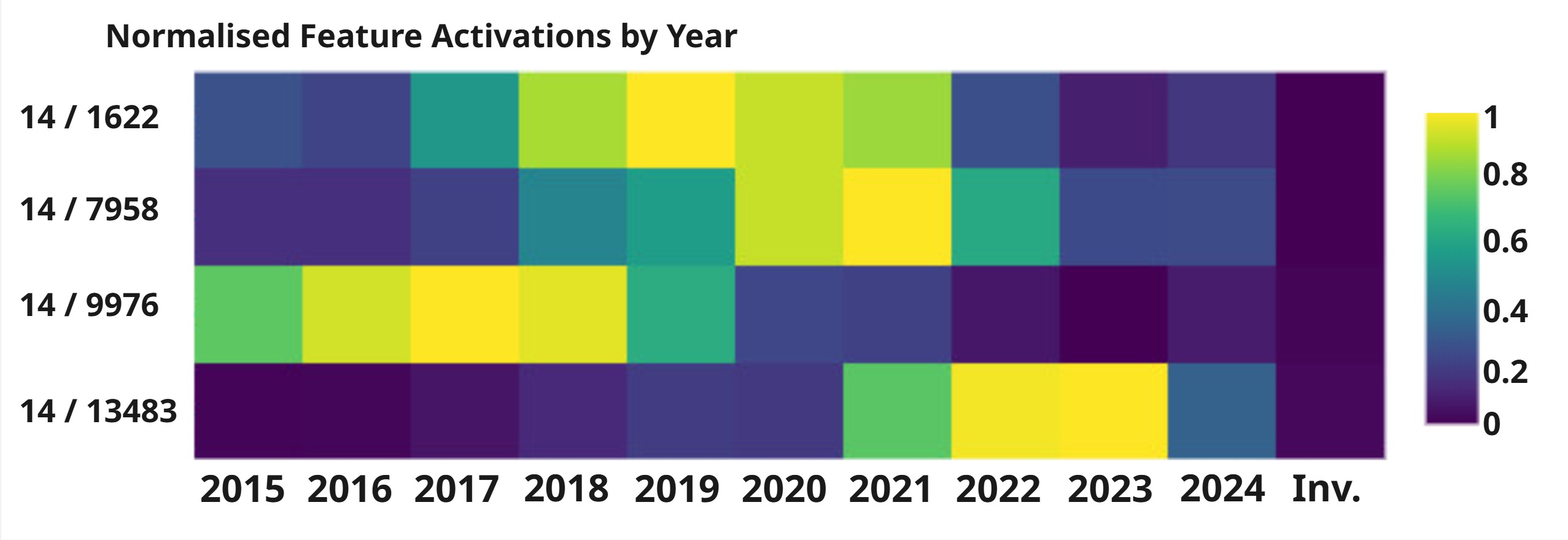}
  \caption{Gemma 2 2B activation heatmap for chrono-semantic features. The graph is normalised for feature ids to highlight the continuous nature of each chrono-semantic feature across a period of time. Notably there are a range of features that cover the temporal range between 2015 up until 2024. Note that the invariant measure is 0 (the last column), demonstrating that the activation is time period specific.}
  \label{fig:Gemma_activation_heatmap_chronosemantic_specialised_intext}
\end{figure}

\textit{ Chrono-semantic Nodes}
From examining the default explanations and node weights across years (Figure \ref{fig:TemporalNodes_FeatureMapping}), we speculate there exist higher temporal representations (\texttt{L14F13483}) attuned to specific years. We test this hypothesis by examining token activations using a set of temporally contextual people, places and events from specific years as defined by Google Gemini 3 (Appendix \ref{app:Activation_Testing_for_Higher_Temporal_Representations}). Surprisingly, we find that tokens related to that time period activate for their corresponding layers (Figure \ref{fig:Gemma_activation_heatmap_chronosemantic_specialised_intext}), whilst tokens that are not specific to that period remain dormant. This suggests that certain subjects (events, places and people) are embedded as temporal markers and could be used as temporal anchors. Qwen3-4b exhibits similar traits, demonstrating specific features that attend to specific moments in time, whilst LLaMA 3.2 1B exhibits a limited relationship.

The parallel, non-hierarchical organisation of temporal features has consequences for intervention strategies (Section \ref{sec:ablation}). If temporal recall depends on simultaneous processing across multiple abstraction levels, single-layer interventions may prove insufficient for precise temporal control. Moreover, the appearance of semantic temporal representations in lower layers and the presence of Chrono-semantic throughout a range of layers (10, 11, 15 and 16 for Qwen3-4b) related to different time frames (2015-2024) complicates intervention strategies.

\textbf{Infuential Chronos Semantic Layers} We specifically highlight this set of representations and create the class because it was clear from inspection that temporal semantic representations (i.e. not numerical or date explanations) were identified in earlier layers, which was unexpected. Our identification of these layers is based on the current transcoder representation and applied vocabulary to the summarisation of the nodes. While these nodes can be reproduced using our methodology, the identification of such nodes is highly dependent on the dynamic neuronpedia source. The interesting and potentially important aspect of this class which can be quantitatively defined is the outlier semantic features which consistently occur in higher layers. To identify the top chrono-semantic influential nodes expressed in this paper as we have developed the following methodology:

\begin{equation}
\mu_y = \operatorname{mean}(w_{\cdot,y}), \quad \sigma_y = \operatorname{sd}(w_{\cdot,y})
\end{equation}

\begin{equation}
\left|w_{f,y} - \mu_y\right| \ge k\sigma_y
\end{equation}

We define this subclass as 2.5 standard deviation from the average weighted activations of the feature within each year. This isolates the nodes () which are aligned with our definition and are expressed as temporal markers in Section \ref{sec:NewtemporalComponents} and captures the year specific semantics, and key semantic layers for common temporal features.

\begin{table*}[ht!]
  \centering
  \resizebox{\textwidth}{!}{%
    \begin{tabular}{l|ccccc|ccc|cc}
         & \multicolumn{5}{c|}{\textbf{F1 Scores}} & \multicolumn{3}{c}{\textbf{RVE Scores}} & \multicolumn{1}{c}{\textbf{Non-Target Ans.}}  \\
         \hline
        \textbf{Year} & \textbf{Inv. F1} & \textbf{Rel. F1} & \textbf{Exp. F1} & \textbf{Temp. Ab. F1} & \textbf{Rand. Ab. F1} & \textbf{Prior rve} & \textbf{Ab. new rve} & \textbf{Rand. new rve} & \textbf{Avg. Logit Diff.}\\
        \hline
        2020 & 0.446 & 0.514 & 0.908 & 0.381 & 0.736 & -0.052 & \textbf{0.035} & -0.048 & 0.025 \\
        2021 & 0.509 & 0.477 & 0.914 & 0.379 & 0.828 & -0.062 & \textbf{0.053} & -0.059 & 0.010 \\
        2022 & 0.390 & 0.425 & 0.917 & 0.302 & 0.813 & -0.051 & \textbf{0.045} & -0.047 & 0.041 \\
        2023 & 0.444 & 0.487 & 0.925 & 0.357 & 0.803 & -0.072 & \textbf{0.042} & -0.059 & 0.049 \\
        2024 & 0.635 & 0.549 & 0.903 & 0.493 & 0.779 & -0.012 & \textbf{0.012} & -0.006 & 0.001 \\
        \hline
        \end{tabular}}
        \caption{Ablation successfully removes temporal grounding, leading degraded F1 due to an increase in non-target (neither relative, invariant nor explicit answers) answer probabilities. Temporal ablation (Temp. Ab. F1) shifts responses toward relative answers (positive Ab. new rve) but achieves F1 scores below even the relative baseline, indicating incomplete factual knowledge. Random ablation maintains baseline rve but reduces F1, confirming our targeted features affect temporal considerations. Gemma 2 2B results on ChronosAlign dataset.}
      \label{tab:table_chqa_100_scores_gemma2_unique}
  \end{table*}

  \begin{table}
  \centering
  \resizebox{\columnwidth}{!}{%
    \begin{tabular}{l|ccc|cc}
              & \multicolumn{3}{c|}{\textbf{F1 Scores}} & \multicolumn{2}{c}{\textbf{RVE Scores}} \\
      \hline
      \textbf{Year} & \textbf{Rel. F1} & \textbf{Exp. F1} & \textbf{Steer F1} &  \textbf{Prior rve} & \textbf{Steer new rve}  \\
      \hline
      2020 & 0.071 & 0.351 & 0.177 & 0.063 & -0.030 \\
      2021 & 0.085 & 0.370 & 0.157 & 0.073 & 0.002 \\
      2022 & 0.020 & 0.416 & 0.151 & 0.069 & 0.013 \\
      2023 & 0.046 & 0.320 & 0.209 & 0.068 & -0.068 \\
      2024 & 0.081 & 0.148 & 0.093 & 0.041 & 0.041 \\
      \hline
      \end{tabular}}
      \caption{F1 and RVE scores for 2020-2024, ChronosAlign dataset, Gemma 2 2b. Using layers above 7, m=5, a=1. Temporal Steering F1 (Steer F1) present F1 scores for each intervention. A score closer to explicit F1 is better. \textit{Prior rve} and \textit{Steer new rve} show probability differences between relative and explicit answers before and after intervention. A smaller \textit{new rve} value relative to \textit{prior rve} value is better}.
    \label{tab:table_chqa_100_scores_gemma2_steering}
  \end{table}

\subsection{Ablation}
\label{sec:ablation}
If temporal representations are compositionally constructed from \commontemporal{} backbone features, ablating these features should substatially reduce explicit temporal grounding across all time periods. Specifically, we speculate that models will revert from explicit answers (e.g., ``In 2024, X won'') to invariant or relative answers (e.g., ``X won'' or ``Who most recently, X won '') when \commontemporal{} features are suppressed. If this hypothesis is correct, the probability difference between invariant or relative and explicit answers (\textit{ive} or \textit{rve}) should invert from negative (favouring explicit) to positive (favouring invariant or relative).

We ablate all 31 \commontemporal{} features (concentrated in layers 4--7) inverting the activation values (Eq. \ref{eq:steering} where $a=1$, $m=-2$). As a control, we perform random ablation of 31 arbitrarily selected features to verify that our identified features form a functionally coherent set rather than spurious correlations. Targeted ablation successfully removes temporal grounding: \textit{Ab. new rve} is positive across all years (e.g., 2020: $-0.052 \rightarrow +0.036$), and \textit{Ab. new ive} approaches 0, but does not turn positive. Table \ref{tab:table_chqa_100_scores_gemma2_unique} confirms the model's shift from explicit to relative answer distributions. Random ablation produces similar \textit{rve} scores to baseline ($-0.050$ vs $-0.052$ for 2020) but with lower F1 scores. This divergence confirms that \commontemporal{} features form a coherent functional unit for temporal grounding, not an artifact of our selection process.

While temporal grounding successfully eliminated temporal considerations, answer quality degrades substantially. F1 scores fall below even the relative baseline (e.g., Temp. Ab. F1 = 0.381 vs. Rel. F1 = 0.514 for 2020)(Table \ref{tab:table_chqa_100_scores_gemma2_unique}). Examining this effect we find that eliminating explicit temporal considerations leads to an increase in probability for non-relative and non-invariant answers, leading to an increase in grounded answer probabilities over a 2010 to 2024 period. These results establish that \commontemporal{} features are necessary for temporal grounding (ablating them removes explicit temporal responses) but are not sufficient for realignment to a relative answer.

\textbf{Higher Temporal Concept Ablation} Having established the \commontemporal{} nodes are fundamental to the accurate recall of time sensitive facts, we further explore how ablating Chrono-semantic representations (specifically features with higher edge weights) affect recall. Notably, ablating these nodes improves F1 results compared to \commontemporal{} node ablation, and causes both the \textit{rve} and \textit{ive} metrics to invert, with invariant and relative answers becoming more probable than explicit ones (Table \ref{tab:table_chqa_100_high_rep_ablation_scores_gemma2}). The improvement of these metrics suggests these specific nodes are vital specialised temporal filters for factual recall.

\textbf{Generalisation across architectures.} We observe similar patterns for LLaMA 3.2 1B and Qwen3-4b (Appendix \ref{app:ablation_Qwen3-4b} \& \ref{app:ablation_LLaMA 3.2 1B}), with architectural variations: LLaMA shows stronger F1 degradation due to compressed temporal representations in lower layers, while Qwen exhibits weaker \textit{rve} inversion (remains negative but shifts significantly from prior). These differences likely reflect Qwen3's substantially lower circuit replacement scores (0.46 vs. 0.73 for Gemma 2 2B) or could be a product of the architectural difference in temporal generation, whereby Qwen has a subtractive definition and Gemma has an additive one (Appendix \ref{app:Activation_Testing_for_Higher_Temporal_Representations}). Furthermore, whilst we are unable to test larger models, we would anticipate that larger models contain more internal space to store specialised MLP temporal components. Where smaller models are required to compact knowledge into few layers and components, larger models have the luxury of more layers and MLP components to dedicate towards specialised representations. Larger models would hopefully provide an easier source to identify temporal features from.

\subsection{Temporal, Topic and Semantic Generalisations}
To test the generalisation of our findings, we investigate the effect of ablation on invariant datasets GSM8K and Commonsense, the ChronosAlign dataset with an altered syntactical structure placing the explicit temporal tokens at the end of the question, and finally the Taqa dataset \citep{Zhao2024-li} with different time periods. Applying interventions on GSM8K and commonsense dataset (Tables: \ref{tab:ablation_temporal_invariant_datasets_gemma_2},\ref{tab:ablation_temporal_invariant_nodes_datasets_qwen3},  \ref{tab:ablation_temporal_invariant_nodes_datasets_llama3}) we find no impact on the overall outcomes for Gemma 2 2B and LLaMA 3.2 1B, but quite a large impact for Qwen3-4b. Testing with semantically different questions, we find that the outcomes are positive across all models, with \textit{new rve} increasing and F1 scores heading towards a relative score. Applying interventions to Taqa, between 2009 and 2024, (Figures: \ref{tab:table_taqa_200_gemma_2}, \ref{tab:table_taqa_200_qwen3},  \ref{tab:table_taqa_200_llama3}) we find very similar outcomes for ablation whereby \textit{new rve} increases relative to the baseline, and the F1 scores revert towards the relative f1 scores. One deviation is with regard to increasing temporal considerations whereby the strength of the ablation diminishes with distance from the 2020-2024 window suggesting that the \commontemporal{} features identified might also have a limited, but larger temporal window.

\subsection{Year-Specific Features Enable Steering}
\label{sec:steering}

For each year 2020--2024, we identify approximately 40 \commontoyear{} features that reliably activate for that year (Section \ref{sec:TemporalFeaturesClusterbyProcessingStage}).  We apply steering by amplifying feature activations ($m = 5$, $a = 1$; Equation \ref{eq:steering}) and measure shifts in explicit vs. relative answer probabilities on ChronosAlign prompts phrased as relative questions. The asymmetry between ablation and steering efficacy suggests that relative and invariant temporal grounding is encoded with greater redundancy; amplifying a subset is insufficient to consistently override the current temporal anchor. Where ablation reliably inverted \textit{rve} signs (Section \ref{sec:ablation}), steering shifts probabilities by 0.002--0.011 across years and only mildly increases F1 scores towards explicit F1 scores. This suggests that there are potentially conflicting underlying temporal considerations, or that the vector representation generated from a static steering method (equation \ref{eq:steering}) is suboptimal to temporally realign the statement.

\textbf{Steering ineffectiveness} Steering was applied in a broad manner with token intervention across the entire prompt excluding the last two tokens (``?'' and ``:'') and a static amplification value ($m$) of 5. Whilst on an individual level our research produces positive results, at a collective perspective our results are suboptimal. This is most likely due to, firstly, the broad application of the intervention leading to a mixed and confused integration of the explicit year. The addition of the year into many token spaces could lead to a confused representation of the year, especially if there are multiple temporal attention components across the token space. Furthermore, where we have amplified explicit temporal representations, we have not removed the relative phrase, potentially leading to a mixed representation of time. Finally, our methodology changes the representation of time, but does change how that information is used. The attention component of the circuit is not accounted for, potentially obscuring the uptake of the new explicit yearly representation. As per the studies conducted by \citet{kamath2025tracing}, future studies could be conducted to investigate how attention vectors affect temporal representation uptake between relative and explicit questions prodiving a clearer indication of optimal steering opportunities.


\subsection{New temporal Components} 
\label{sec:NewtemporalComponents}
The identification of higher temporal representations at layers 14 and above is at odds with findings from \citet{Park2025-ok} and \citet{govindan-etal-2025-temporal}. In the case of \citet{Park2025-ok}, applying their methodology to Gemma 2 2B (Appendix: \ref{app:appendix_eapig_graphs}), we note temporal attention nodes located at lower layers (0, 2 and 7). \citet{govindan-etal-2025-temporal} confirms this finding presenting experiments which highlight how layers lower than 12 are large contributors to steering models towards explicit temporal outcomes. Our findings identify a new set of temporal representations at layer 14 which has the ability to negate explicit answers and a strong potential to realign temporal representations. 


\textbf{Additive vs Subtractive Temporal Representations} 
The creation of higher conceptual temporal representations and their ability to ablate and steer temporal concepts suggests that these semantic representations act as filters to the preceding question text. We, however, note a stark contrast between the development of temporal representations of Qwen3-4b and Gemma 2 2B. Qwen3-4b contains large negatively weighted chrono-semantic nodes (\texttt{L16H45664}, \texttt{L15H127839}, \texttt{L11H148703}, \texttt{L10H125625}) and Gemma 2 2B contains large positively weighted nodes (\texttt{L14H13483}, \texttt{L14H1622}, \texttt{L14H9976}). This suggests Qwen3-4b is subtractive in its generation of time, representing all periods, and ablating unrequired temporal definitions as a last step, while Gemma 2 2B undertakes the reverse.

\section{Related Studies}
\subsection{Temporal Knowledge and Alignment}
Various studies have aimed to understand factual recall and consistency within LLMs, aimed at untangling recall for subject, relationship, object (SRO) triplets \citep{Mitchell2022-nf,Li2024-hc,Hartvigsen2023-wj,Mitchell2022-ay, wang-etal-2024-unveiling, yu-ananiadou-2024-neuron}. These studies demonstrate that internal representations for factual recall are malleable and dynamically distributed across layers \citep{Geva2023-le}. Temporal reasoning has received separate attention\citep{wang-zhao-2024-tram, Son2023-gv, gruber-etal-2025-complextempqa}, though this work is largely concerned with multi-hop inference over time rather than the representation of time itself during recall. Efforts to correct temporal misalignment have focused on external interventions, such as fine-tuning, RAG and activation engineering \citep{Zhao2024-li, Mousavi2024-wm,govindan-etal-2025-temporal}. These investigations do not examine internal mechanisms that give rise to misalignment in the first place. \citet{Park2025-ok} is the closest prior work to ours, identifying temporal attention heads via EAP-IG circuit analysis. We extend this by moving to the MLP feature level via transcoders, uncovering representations that attention focussed analysis cannot resolve, and identifying a new class of higher layer \textit{chrono-semantic} components absent from prior accounts.

\subsection{Sparse Autoencoders}

Building on these insights, Sparse autoencoders (SAEs) provide complementary tools for temporal representation analysis. SAEs represent a highly effective tool for uncovering and probing polysemantic representations within the internal layers of LLMs \citep{shu-etal-2025-survey}. Recent studies have employed them to investigate bias reduction \citep{cyberey-etal-2025-unsupervised}, artificial text detection \citep{kuznetsov-etal-2025-feature}, and factual recall for static, non-temporal information \citep{wang-etal-2024-unveiling}. Supporting these exploratory studies are a range of research focussed on generating improved and efficient representations \citep{bhalla2025temporalsparseautoencodersleveraging, ghilardi-etal-2025-group}. Our research presents a new perspective, exploring temporal representations for time sensitive factual recall across several models using per layer transcoders.

\section{Conclusion}
We methodically investigate how LLMs use MLP layers to represent time and recover time sensitive information. We identify three sets of MLP features: (1) \commontemporal{}, encoding numerical time representations, forming the backbone of temporal recall; (2) \commontoyear{}, encoding temporally specific indicators that act as a temporal filter for specific representations of a year; (3) \textit{chrono-semantic} a subset of (2) that are powerful temporal nodes tied to specific points in time.

Where prior studies have focussed on components such as attention or MLP nodes, our work establishes a foundational understanding of why MLP components activate in time sensitive recall scenarios. We locate new temporal components, such as layer 14 MLP in Gemma 2, and identify internal temporal representations that can be manipulated to perturb or assist time sensitive recall. By characterising these features, we lay the groundwork for developing targeted interventions for encoding and retrieving time-sensitive factual knowledge within LLMs.

\section{Limitations}
Our approach demonstrates promising results, identifying temporal nodes that have temporal ablative and steering capability across a range of datasets. However, we acknowledge some limitations in our research. First, we are limited in our testing of models that have available transcoder models. We are unable to test this on larger LLMs due to both transcoder limitations and computational constraints.

\textbf{MLP transcoders} Our studies focussed on MLP components of LLMs. We are identifying what information moved, but are limited in understanding why it moved. The 'why' component is associated with the attention mechanisms which as discussed in section \ref{sec:steering}, could have a strong influence on steering outcomes.

\textbf{Optimisation} Whilst we identified a strong set of results using a limited grid search for our range of steering parameters, we have not exhaustively searched for global optimal parameters for ablation or steering, which could be achieved through learned scaled representations of temporal features.

\textbf{Transcoder Feature Interpretation} Our research utilises existing transcoder explanations where available and creates one for LLaMA 3.2 1B, where no explanation set existed. Explainability and interpretation of transcoder features is an active area of research \citep{puri-etal-2025-fade, cho-etal-2025-faithfulsae} and whilst we note that there exists complex temporal representations throughout the layers of a model, and have validated our conceptual temporal findings with token activation tests, validating the accuracy of these explanations is beyond the scope of this work and could form part of future research.

\textbf{Temporal Representational Filters} We demonstrate that there exists a range of temporal anchors from mid to higher layers in a model. Our research suggests that these chrono-semantic representations encapsulate information relevant to that period of time. Although our findings identify a range of temporal anchors that encapsulate time specific information, the relationship between these representations and the rest of the prompt is likely multifaceted. The non-trivial coupling of temporal signals with domain specificity, answer phrasing and syntax, and knowledge retrieval suggests that temporal alignment does not occur in isolation. Mapping these interactions remains an essential direction for future study to ensure robust performance across diverse query types.

\bibliography{custom}

@inproceedings{Mitchell2022-nf,
  title     = {Fast Model Editing at Scale},
  author    = {Mitchell, Eric and Lin, Charles and Bosselut, Antoine and Finn,
               Chelsea and Manning, Christopher D},
  booktitle = {International Conference on Learning Representations},
  year      = 2022
}

@inproceedings{govindan-etal-2025-temporal,
  title     = {Temporal Alignment of Time Sensitive Facts with Activation Engineering},
  author    = {Govindan, Sanjay  and
               Pagnucco, Maurice  and
               Song, Yang},
  editor    = {Christodoulopoulos, Christos  and
               Chakraborty, Tanmoy  and
               Rose, Carolyn  and
               Peng, Violet},
  booktitle = {Findings of the Association for Computational Linguistics: EMNLP 2025},
  month     = nov,
  year      = {2025},
  address   = {Suzhou, China},
  publisher = {Association for Computational Linguistics},
  url       = {https://aclanthology.org/2025.findings-emnlp.404/},
  doi       = {10.18653/v1/2025.findings-emnlp.404},
  pages     = {7640--7657},
  isbn      = {979-8-89176-335-7}
}

@inproceedings{Park2025-ok,
  title     = {Does time have its place? Temporal heads: Where language models
               recall time-specific information},
  author    = {Park, Yein and Yoon, Chanwoong and Park, Jungwoo and Jeong,
               Minbyul and Kang, Jaewoo},
  editor    = {Che, Wanxiang and Nabende, Joyce and Shutova, Ekaterina and
               Pilehvar, Mohammad Taher},
  booktitle = {Proceedings of the 63rd Annual Meeting of the Association for
               Computational Linguistics (Volume 1: Long Papers)},
  publisher = {Association for Computational Linguistics},
  address   = {Stroudsburg, PA, USA},
  pages     = {16616--16643},
  year      = 2025
}

@inproceedings{Luu2022-fk,
  title     = {Time Waits for No One! Analysis and Challenges of Temporal
               Misalignment},
  author    = {Luu, Kelvin and Khashabi, Daniel and Gururangan, Suchin and
               Mandyam, Karishma and Smith, Noah A},
  editor    = {Carpuat, Marine and de Marneffe, Marie-Catherine and Meza Ruiz,
               Ivan Vladimir},
  booktitle = {Proceedings of the 2022 Conference of the North American Chapter
               of the Association for Computational Linguistics: Human Language
               Technologies},
  publisher = {Association for Computational Linguistics},
  address   = {Seattle, United States},
  pages     = {5944--5958},
  month     = jul,
  year      = 2022
}

@misc{lieberum2024gemmascopeopensparse,
  title         = {Gemma Scope: Open Sparse Autoencoders Everywhere All At Once on Gemma 2},
  author        = {Tom Lieberum and Senthooran Rajamanoharan and Arthur Conmy and Lewis Smith and Nicolas Sonnerat and Vikrant Varma and János Kramár and Anca Dragan and Rohin Shah and Neel Nanda},
  year          = {2024},
  eprint        = {2408.05147},
  archiveprefix = {arXiv},
  primaryclass  = {cs.LG},
  url           = {https://arxiv.org/abs/2408.05147}
}

@misc{gemma2024,
  title        = {Gemma 2: Improving Open Language Models at a Practical Size},
  author       = {{Google DeepMind} and et al.},
  year         = {2024},
  month        = jul,
  institution  = {{Google DeepMind}},
  howpublished = {arXiv:2408.00118},
  link         = {https://arxiv.org/pdf/2408.00118.pdf}
}

@inproceedings{cho-etal-2025-faithfulsae,
  title     = {{F}aithful{SAE}: Towards Capturing Faithful Features with Sparse Autoencoders without External Datasets Dependency},
  author    = {Cho, Seonglae  and
               Oh, Harryn  and
               Lee, Donghyun  and
               Vieira, Luis Rodrigues  and
               Bermingham, Andrew  and
               Sayed, Ziad El},
  editor    = {Zhao, Jin  and
               Wang, Mingyang  and
               Liu, Zhu},
  booktitle = {Proceedings of the 63rd Annual Meeting of the Association for Computational Linguistics (Volume 4: Student Research Workshop)},
  month     = jul,
  year      = {2025},
  address   = {Vienna, Austria},
  publisher = {Association for Computational Linguistics},
  url       = {https://aclanthology.org/2025.acl-srw.20/},
  doi       = {10.18653/v1/2025.acl-srw.20},
  pages     = {297--314},
  isbn      = {979-8-89176-254-1}
}

@inproceedings{puri-etal-2025-fade,
  title     = {{FADE}: Why Bad Descriptions Happen to Good Features},
  author    = {Puri, Bruno  and
               Jain, Aakriti  and
               Golimblevskaia, Elena  and
               Kahardipraja, Patrick  and
               Wiegand, Thomas  and
               Samek, Wojciech  and
               Lapuschkin, Sebastian},
  editor    = {Che, Wanxiang  and
               Nabende, Joyce  and
               Shutova, Ekaterina  and
               Pilehvar, Mohammad Taher},
  booktitle = {Findings of the Association for Computational Linguistics: ACL 2025},
  month     = jul,
  year      = {2025},
  address   = {Vienna, Austria},
  publisher = {Association for Computational Linguistics},
  url       = {https://aclanthology.org/2025.findings-acl.881/},
  doi       = {10.18653/v1/2025.findings-acl.881},
  pages     = {17138--17160},
  isbn      = {979-8-89176-256-5}
}

@inproceedings{liu-niehues-2025-conditions,
  title     = {Conditions for Catastrophic Forgetting in Multilingual Translation},
  author    = {Liu, Danni  and
               Niehues, Jan},
  editor    = {Adelani, David Ifeoluwa  and
               Arnett, Catherine  and
               Ataman, Duygu  and
               Chang, Tyler A.  and
               Gonen, Hila  and
               Raja, Rahul  and
               Schmidt, Fabian  and
               Stap, David  and
               Wang, Jiayi},
  booktitle = {Proceedings of the 5th Workshop on Multilingual Representation Learning (MRL 2025)},
  month     = nov,
  year      = {2025},
  address   = {Suzhuo, China},
  publisher = {Association for Computational Linguistics},
  url       = {https://aclanthology.org/2025.mrl-main.23/},
  doi       = {10.18653/v1/2025.mrl-main.23},
  pages     = {347--359},
  isbn      = {979-8-89176-345-6}
}

@misc{qwen3technicalreport,
  title         = {Qwen3 Technical Report},
  author        = {Qwen Team},
  year          = {2025},
  eprint        = {2505.09388},
  archiveprefix = {arXiv},
  primaryclass  = {cs.CL},
  url           = {https://arxiv.org/abs/2505.09388}
}

@article{doi:10.1073/pnas.1611835114,
  author   = {James Kirkpatrick  and Razvan Pascanu  and Neil Rabinowitz  and Joel Veness  and Guillaume Desjardins  and Andrei A. Rusu  and Kieran Milan  and John Quan  and Tiago Ramalho  and Agnieszka Grabska-Barwinska  and Demis Hassabis  and Claudia Clopath  and Dharshan Kumaran  and Raia Hadsell },
  title    = {Overcoming catastrophic forgetting in neural networks},
  journal  = {Proceedings of the National Academy of Sciences},
  volume   = {114},
  number   = {13},
  pages    = {3521-3526},
  year     = {2017},
  doi      = {10.1073/pnas.1611835114},
  url      = {https://www.pnas.org/doi/abs/10.1073/pnas.1611835114},
  eprint   = {https://www.pnas.org/doi/pdf/10.1073/pnas.1611835114}
}

@inproceedings{li-etal-2024-revisiting,
  title     = {Revisiting Catastrophic Forgetting in Large Language Model Tuning},
  author    = {Li, Hongyu  and
               Ding, Liang  and
               Fang, Meng  and
               Tao, Dacheng},
  editor    = {Al-Onaizan, Yaser  and
               Bansal, Mohit  and
               Chen, Yun-Nung},
  booktitle = {Findings of the Association for Computational Linguistics: EMNLP 2024},
  month     = nov,
  year      = {2024},
  address   = {Miami, Florida, USA},
  publisher = {Association for Computational Linguistics},
  url       = {https://aclanthology.org/2024.findings-emnlp.249/},
  doi       = {10.18653/v1/2024.findings-emnlp.249},
  pages     = {4297--4308}
}

@proceedings{blackboxnlp-ws-2025-1,
  title     = {Proceedings of the 8th BlackboxNLP Workshop: Analyzing and Interpreting Neural Networks for NLP},
  editor    = {Belinkov, Yonatan  and
               Mueller, Aaron  and
               Kim, Najoung  and
               Mohebbi, Hosein  and
               Chen, Hanjie  and
               Arad, Dana  and
               Sarti, Gabriele},
  month     = nov,
  year      = {2025},
  address   = {Suzhou, China},
  publisher = {Association for Computational Linguistics},
  url       = {https://aclanthology.org/2025.blackboxnlp-1.0/},
  doi       = {10.18653/v1/2025.blackboxnlp-1.0},
  isbn      = {979-8-89176-346-3}
}

@article{Grattafiori2024-nw,
  title         = {The Llama 3 herd of models},
  author        = {Dubey, Abhimanyu and Jauhri, Abhinav
                   and Pandey, Abhinav and Kadian, Abhishek and Al-Dahle, Ahmad
                   and Letman, Aiesha and Mathur, Akhil and Schelten, Alan and
                   Vaughan, Alex and Yang, Amy and Fan, Angela and more},
  journal       = {arXiv [cs.AI]},
  month         = jul,
  year          = 2024,
  archiveprefix = {arXiv},
  primaryclass  = {cs.AI}
}

@inproceedings{Hernandez2024-hx,
  title     = {Inspecting and Editing Knowledge Representations in Language
               Models},
  author    = {Hernandez, Evan and Li, Belinda Z and Andreas, Jacob},
  booktitle = {First Conference on Language Modeling},
  year      = 2024
}

@inproceedings{Geva2023-le,
  title     = {Dissecting Recall of Factual Associations in Auto-Regressive
               Language Models},
  author    = {Geva, Mor and Bastings, Jasmijn and Filippova, Katja and
               Globerson, Amir},
  editor    = {Bouamor, Houda and Pino, Juan and Bali, Kalika},
  booktitle = {Proceedings of the 2023 Conference on Empirical Methods in
               Natural Language Processing},
  publisher = {Association for Computational Linguistics},
  address   = {Singapore},
  pages     = {12216--12235},
  month     = dec,
  year      = 2023
}

@misc{circuittracer,
  author       = {Hanna, Michael and Piotrowski, Mateusz and Lindsey, Jack and Ameisen, Emmanuel},
  title        = {circuit-tracer},
  howpublished = {\url{https://github.com/safety-research/circuit-tracer}},
  note         = {The first two authors contributed equally and are listed alphabetically.},
  year         = {2025}
}

@article{cobbe2021gsm8k,
  title   = {Training Verifiers to Solve Math Word Problems},
  author  = {Cobbe, Karl and Kosaraju, Vineet and Bavarian, Mohammad and Chen, Mark and Jun, Heewoo and Kaiser, Lukasz and Plappert, Matthias and Tworek, Jerry and Hilton, Jacob and Nakano, Reiichiro and Hesse, Christopher and Schulman, John},
  journal = {arXiv preprint arXiv:2110.14168},
  year    = {2021}
}

@article{ameisen2025circuit,
  author  = {Ameisen, Emmanuel and Lindsey, Jack and Pearce, Adam and Gurnee, Wes and Turner, Nicholas L. and Chen, Brian and Citro, Craig and Abrahams, David and Carter, Shan and Hosmer, Basil and Marcus, Jonathan and Sklar, Michael and Templeton, Adly and Bricken, Trenton and McDougall, Callum and Cunningham, Hoagy and Henighan, Thomas and Jermyn, Adam and Jones, Andy and Persic, Andrew and Qi, Zhenyi and Ben Thompson, T. and Zimmerman, Sam and Rivoire, Kelley and Conerly, Thomas and Olah, Chris and Batson, Joshua},
  title   = {Circuit Tracing: Revealing Computational Graphs in Language Models},
  journal = {Transformer Circuits Thread},
  year    = {2025},
  url     = {https://transformer-circuits.pub/2025/attribution-graphs/methods.html}
}

@article{DBLP:journals/corr/abs-2405-17969,
  author     = {Yunzhi Yao and
                Ningyu Zhang and
                Zekun Xi and
                Mengru Wang and
                Ziwen Xu and
                Shumin Deng and
                Huajun Chen},
  title      = {Knowledge Circuits in Pretrained Transformers},
  journal    = {CoRR},
  volume     = {abs/2405.17969},
  year       = {2024},
  url        = {https://doi.org/10.48550/arXiv.2405.17969},
  doi        = {10.48550/ARXIV.2405.17969},
  eprinttype = {arXiv},
  eprint     = {2405.17969},
  bibsource  = {dblp computer science bibliography, https://dblp.org}
}

@inproceedings{Zhang2025-gq,
  title     = {{MRAG}: A modular retrieval framework for time-sensitive question
               answering},
  author    = {Zhang, Siyue and Xue, Yuxiang and Zhang, Yiming and Wu, Xiaobao
               and Luu, Anh Tuan and Zhao, Chen},
  booktitle = {Findings of the Association for Computational Linguistics: EMNLP
               2025},
  publisher = {Association for Computational Linguistics},
  address   = {Stroudsburg, PA, USA},
  volume    = 2025,
  pages     = {3080--3118},
  year      = 2025,
  language  = {en}
}

@inproceedings{Zhao2024-li,
  title     = {Set the Clock: Temporal Alignment of Pretrained Language Models},
  author    = {Zhao, Bowen and Brumbaugh, Zander and Wang, Yizhong and
               Hajishirzi, Hannaneh and Smith, Noah},
  editor    = {Ku, Lun-Wei and Martins, Andre and Srikumar, Vivek},
  booktitle = {Findings of the Association for Computational Linguistics: ACL
               2024},
  publisher = {Association for Computational Linguistics},
  address   = {Bangkok, Thailand},
  pages     = {15015--15040},
  month     = aug,
  year      = 2024
}

@inproceedings{Mousavi2024-wm,
  title     = {{DyKnow}: Dynamically verifying time-sensitive factual knowledge
               in {LLMs}},
  author    = {Mousavi, Seyed Mahed and Alghisi, Simone and Riccardi, Giuseppe},
  editor    = {Al-Onaizan, Yaser and Bansal, Mohit and Chen, Yun-Nung},
  booktitle = {Findings of the Association for Computational Linguistics: EMNLP
               2024},
  publisher = {Association for Computational Linguistics},
  address   = {Stroudsburg, PA, USA},
  pages     = {8014--8029},
  year      = 2024
}

@misc{bhalla2025temporalsparseautoencodersleveraging,
  title         = {Temporal Sparse Autoencoders: Leveraging the Sequential Nature of Language for Interpretability},
  author        = {Usha Bhalla and Alex Oesterling and Claudio Mayrink Verdun and Himabindu Lakkaraju and Flavio P. Calmon},
  year          = {2025},
  eprint        = {2511.05541},
  archiveprefix = {arXiv},
  primaryclass  = {cs.CL},
  url           = {https://arxiv.org/abs/2511.05541}
}

@inproceedings{NEURIPS2024_2b8f4db0,
  author    = {Dunefsky, Jacob and Chlenski, Philippe and Nanda, Neel},
  booktitle = {Advances in Neural Information Processing Systems},
  doi       = {10.52202/079017-0768},
  editor    = {A. Globerson and L. Mackey and D. Belgrave and A. Fan and U. Paquet and J. Tomczak and C. Zhang},
  pages     = {24375--24410},
  publisher = {Curran Associates, Inc.},
  title     = {Transcoders find interpretable LLM feature circuits},
  url       = {https://proceedings.neurips.cc/paper_files/paper/2024/file/2b8f4db0464cc5b6e9d5e6bea4b9f308-Paper-Conference.pdf},
  volume    = {37},
  year      = {2024}
}

@inproceedings{yu-ananiadou-2024-neuron,
  title     = {Neuron-Level Knowledge Attribution in Large Language Models},
  author    = {Yu, Zeping  and
               Ananiadou, Sophia},
  editor    = {Al-Onaizan, Yaser  and
               Bansal, Mohit  and
               Chen, Yun-Nung},
  booktitle = {Proceedings of the 2024 Conference on Empirical Methods in Natural Language Processing},
  month     = nov,
  year      = {2024},
  address   = {Miami, Florida, USA},
  publisher = {Association for Computational Linguistics},
  url       = {https://aclanthology.org/2024.emnlp-main.191/},
  doi       = {10.18653/v1/2024.emnlp-main.191},
  pages     = {3267--3280}
}

@inproceedings{10.5555/3692070.3692115,
  author    = {Allen-Zhu, Zeyuan and Li, Yuanzhi},
  title     = {Physics of language models: part 3.1, knowledge storage and extraction},
  year      = {2024},
  publisher = {JMLR.org},
  booktitle = {Proceedings of the 41st International Conference on Machine Learning},
  articleno = {45},
  numpages  = {11},
  location  = {Vienna, Austria},
  series    = {ICML'24}
}

@inproceedings{vu-etal-2024-freshllms,
  title     = {{F}resh{LLM}s: Refreshing Large Language Models with Search Engine Augmentation},
  author    = {Vu, Tu  and
               Iyyer, Mohit  and
               Wang, Xuezhi  and
               Constant, Noah  and
               Wei, Jerry  and
               Wei, Jason  and
               Tar, Chris  and
               Sung, Yun-Hsuan  and
               Zhou, Denny  and
               Le, Quoc  and
               Luong, Thang},
  editor    = {Ku, Lun-Wei  and
               Martins, Andre  and
               Srikumar, Vivek},
  booktitle = {Findings of the Association for Computational Linguistics: ACL 2024},
  month     = aug,
  year      = {2024},
  address   = {Bangkok, Thailand},
  publisher = {Association for Computational Linguistics},
  url       = {https://aclanthology.org/2024.findings-acl.813/},
  doi       = {10.18653/v1/2024.findings-acl.813},
  pages     = {13697--13720}
}

@inproceedings{ou-etal-2025-llms,
  title     = {How Do {LLM}s Acquire New Knowledge? A Knowledge Circuits Perspective on Continual Pre-Training},
  author    = {Ou, Yixin  and
               Yao, Yunzhi  and
               Zhang, Ningyu  and
               Jin, Hui  and
               Sun, Jiacheng  and
               Deng, Shumin  and
               Li, Zhenguo  and
               Chen, Huajun},
  editor    = {Che, Wanxiang  and
               Nabende, Joyce  and
               Shutova, Ekaterina  and
               Pilehvar, Mohammad Taher},
  booktitle = {Findings of the Association for Computational Linguistics: ACL 2025},
  month     = jul,
  year      = {2025},
  address   = {Vienna, Austria},
  publisher = {Association for Computational Linguistics},
  url       = {https://aclanthology.org/2025.findings-acl.1021/},
  doi       = {10.18653/v1/2025.findings-acl.1021},
  pages     = {19889--19913},
  isbn      = {979-8-89176-256-5}
}

@inproceedings{cho-etal-2025-understanding,
  title     = {Understanding Token Probability Encoding in Output Embeddings},
  author    = {Cho, Hakaze  and
               Sakai, Yoshihiro  and
               Tanaka, Kenshiro  and
               Kato, Mariko  and
               Inoue, Naoya},
  editor    = {Rambow, Owen  and
               Wanner, Leo  and
               Apidianaki, Marianna  and
               Al-Khalifa, Hend  and
               Eugenio, Barbara Di  and
               Schockaert, Steven},
  booktitle = {Proceedings of the 31st International Conference on Computational Linguistics},
  month     = jan,
  year      = {2025},
  address   = {Abu Dhabi, UAE},
  publisher = {Association for Computational Linguistics},
  url       = {https://aclanthology.org/2025.coling-main.708/},
  pages     = {10618--10633}
}

@inproceedings{pham-etal-2024-whos,
  title     = {Who{'}s Who: Large Language Models Meet Knowledge Conflicts in Practice},
  author    = {Pham, Quang Hieu  and
               Ngo, Hoang  and
               Luu, Anh Tuan  and
               Nguyen, Dat Quoc},
  editor    = {Al-Onaizan, Yaser  and
               Bansal, Mohit  and
               Chen, Yun-Nung},
  booktitle = {Findings of the Association for Computational Linguistics: EMNLP 2024},
  month     = nov,
  year      = {2024},
  address   = {Miami, Florida, USA},
  publisher = {Association for Computational Linguistics},
  url       = {https://aclanthology.org/2024.findings-emnlp.593/},
  doi       = {10.18653/v1/2024.findings-emnlp.593},
  pages     = {10142--10151}
}

@inproceedings{jiang2025time,
  title     = {Time Waits for No Benchmark: Exploring the Temporal Misalignment between Static Benchmarks, Modern {LLM}s, and the Real World},
  author    = {Xunyi Jiang and Dingyi Chang and Xin Xu},
  booktitle = {NeurIPS 2025 Workshop on Evaluating the Evolving LLM Lifecycle: Benchmarks, Emergent Abilities, and Scaling},
  year      = {2025},
  url       = {https://openreview.net/forum?id=WRMjVoJfCY}
}

@article{lindsey2025biology,
  author  = {Lindsey, Jack and Gurnee, Wes and Ameisen, Emmanuel and Chen, Brian and Pearce, Adam and Turner, Nicholas L. and Citro, Craig and Abrahams, David and Carter, Shan and Hosmer, Basil and Marcus, Jonathan and Sklar, Michael and Templeton, Adly and Bricken, Trenton and McDougall, Callum and Cunningham, Hoagy and Henighan, Thomas and Jermyn, Adam and Jones, Andy and Persic, Andrew and Qi, Zhenyi and Thompson, T. Ben and Zimmerman, Sam and Rivoire, Kelley and Conerly, Thomas and Olah, Chris and Batson, Joshua},
  title   = {On the Biology of a Large Language Model},
  journal = {Transformer Circuits Thread},
  year    = {2025},
  url     = {https://transformer-circuits.pub/2025/attribution-graphs/biology.html}
}

@inproceedings{marks2025sparse,
  title     = {Sparse Feature Circuits: Discovering and Editing Interpretable Causal Graphs in Language Models},
  author    = {Samuel Marks and Can Rager and Eric J Michaud and Yonatan Belinkov and David Bau and Aaron Mueller},
  booktitle = {The Thirteenth International Conference on Learning Representations},
  year      = {2025},
  url       = {https://openreview.net/forum?id=I4e82CIDxv}
}

@inproceedings{ghilardi-etal-2025-group,
  title     = {Group-{SAE}: Efficient Training of Sparse Autoencoders for Large Language Models via Layer Groups},
  author    = {Ghilardi, Davide  and
               Belotti, Federico  and
               Molinari, Marco  and
               Ma, Tao  and
               Palmonari, Matteo},
  editor    = {Christodoulopoulos, Christos  and
               Chakraborty, Tanmoy  and
               Rose, Carolyn  and
               Peng, Violet},
  booktitle = {Proceedings of the 2025 Conference on Empirical Methods in Natural Language Processing},
  month     = nov,
  year      = {2025},
  address   = {Suzhou, China},
  publisher = {Association for Computational Linguistics},
  url       = {https://aclanthology.org/2025.emnlp-main.942/},
  doi       = {10.18653/v1/2025.emnlp-main.942},
  pages     = {18657--18677},
  isbn      = {979-8-89176-332-6}
}

@inproceedings{shu-etal-2025-survey,
  title     = {A Survey on Sparse Autoencoders: Interpreting the Internal Mechanisms of Large Language Models},
  author    = {Shu, Dong  and
               Wu, Xuansheng  and
               Zhao, Haiyan  and
               Rai, Daking  and
               Yao, Ziyu  and
               Liu, Ninghao  and
               Du, Mengnan},
  editor    = {Christodoulopoulos, Christos  and
               Chakraborty, Tanmoy  and
               Rose, Carolyn  and
               Peng, Violet},
  booktitle = {Findings of the Association for Computational Linguistics: EMNLP 2025},
  month     = nov,
  year      = {2025},
  address   = {Suzhou, China},
  publisher = {Association for Computational Linguistics},
  url       = {https://aclanthology.org/2025.findings-emnlp.89/},
  doi       = {10.18653/v1/2025.findings-emnlp.89},
  pages     = {1690--1712},
  isbn      = {979-8-89176-335-7}
}

@inproceedings{kuznetsov-etal-2025-feature,
  title     = {Feature-Level Insights into Artificial Text Detection with Sparse Autoencoders},
  author    = {Kuznetsov, Kristian  and
               Kushnareva, Laida  and
               Razzhigaev, Anton  and
               Druzhinina, Polina  and
               Voznyuk, Anastasia  and
               Piontkovskaya, Irina  and
               Burnaev, Evgeny  and
               Barannikov, Serguei},
  editor    = {Che, Wanxiang  and
               Nabende, Joyce  and
               Shutova, Ekaterina  and
               Pilehvar, Mohammad Taher},
  booktitle = {Findings of the Association for Computational Linguistics: ACL 2025},
  month     = jul,
  year      = {2025},
  address   = {Vienna, Austria},
  publisher = {Association for Computational Linguistics},
  url       = {https://aclanthology.org/2025.findings-acl.1321/},
  doi       = {10.18653/v1/2025.findings-acl.1321},
  pages     = {25727--25748},
  isbn      = {979-8-89176-256-5}
}

@inproceedings{cyberey-etal-2025-unsupervised,
  title     = {Unsupervised Concept Vector Extraction for Bias Control in {LLM}s},
  author    = {Cyberey, Hannah  and
               Ji, Yangfeng  and
               Evans, David},
  editor    = {Christodoulopoulos, Christos  and
               Chakraborty, Tanmoy  and
               Rose, Carolyn  and
               Peng, Violet},
  booktitle = {Proceedings of the 2025 Conference on Empirical Methods in Natural Language Processing},
  month     = nov,
  year      = {2025},
  address   = {Suzhou, China},
  publisher = {Association for Computational Linguistics},
  url       = {https://aclanthology.org/2025.emnlp-main.1439/},
  doi       = {10.18653/v1/2025.emnlp-main.1439},
  pages     = {28333--28355},
  isbn      = {979-8-89176-332-6}
}

@inproceedings{wang-etal-2024-unveiling,
  title     = {Unveiling Factual Recall Behaviors of Large Language Models through Knowledge Neurons},
  author    = {Wang, Yifei  and
               Chen, Yuheng  and
               Wen, Wanting  and
               Sheng, Yu  and
               Li, Linjing  and
               Zeng, Daniel Dajun},
  editor    = {Al-Onaizan, Yaser  and
               Bansal, Mohit  and
               Chen, Yun-Nung},
  booktitle = {Proceedings of the 2024 Conference on Empirical Methods in Natural Language Processing},
  month     = nov,
  year      = {2024},
  address   = {Miami, Florida, USA},
  publisher = {Association for Computational Linguistics},
  url       = {https://aclanthology.org/2024.emnlp-main.420/},
  doi       = {10.18653/v1/2024.emnlp-main.420},
  pages     = {7388--7402}
}

@inproceedings{gruber-etal-2025-complextempqa,
  title     = {\{C\}omplex\{T\}emp\{QA\}: A 100m Dataset for Complex Temporal Question Answering},
  author    = {Gruber, Raphael  and
               Abdallah, Abdelrahman  and
               F{\"a}rber, Michael  and
               Jatowt, Adam},
  editor    = {Christodoulopoulos, Christos  and
               Chakraborty, Tanmoy  and
               Rose, Carolyn  and
               Peng, Violet},
  booktitle = {Proceedings of the 2025 Conference on Empirical Methods in Natural Language Processing},
  month     = nov,
  year      = {2025},
  address   = {Suzhou, China},
  publisher = {Association for Computational Linguistics},
  url       = {https://aclanthology.org/2025.emnlp-main.463/},
  doi       = {10.18653/v1/2025.emnlp-main.463},
  pages     = {9111--9123},
  isbn      = {979-8-89176-332-6}
}

@inproceedings{Son2023-gv,
  title     = {Time-aware representation learning for time-sensitive question
               answering},
  author    = {Son, Jungbin and Oh, Alice},
  booktitle = {Findings of the Association for Computational Linguistics: EMNLP
               2023},
  publisher = {Association for Computational Linguistics},
  address   = {Stroudsburg, PA, USA},
  pages     = {70--77},
  month     = dec,
  year      = 2023
}

@inproceedings{wang-zhao-2024-tram,
  title     = {{TRAM}: Benchmarking Temporal Reasoning for Large Language Models},
  author    = {Wang, Yuqing  and
               Zhao, Yun},
  editor    = {Ku, Lun-Wei  and
               Martins, Andre  and
               Srikumar, Vivek},
  booktitle = {Findings of the Association for Computational Linguistics: ACL 2024},
  month     = aug,
  year      = {2024},
  address   = {Bangkok, Thailand},
  publisher = {Association for Computational Linguistics},
  url       = {https://aclanthology.org/2024.findings-acl.382/},
  doi       = {10.18653/v1/2024.findings-acl.382},
  pages     = {6389--6415}
}

@article{Mitchell2022-ay,
  title         = {Memory-based model editing at scale},
  author        = {Mitchell, E and Lin, Charles and Bosselut, Antoine and
                   Manning, Christopher D and Finn, Chelsea},
  journal       = {ICML},
  volume        = {abs/2206.06520},
  month         = jun,
  year          = 2022,
  archiveprefix = {arXiv},
  primaryclass  = {cs.AI}
}

@inproceedings{Lewis2020-wi,
  title     = {Retrieval-Augmented Generation for Knowledge-Intensive {NLP}
               Tasks},
  author    = {Lewis, Patrick and Perez, Ethan and Piktus, Aleksandra and
               Petroni, Fabio and Karpukhin, Vladimir and Goyal, Naman and
               Küttler, Heinrich and Lewis, Mike and Yih, Wen-Tau and
               Rocktäschel, Tim and Riedel, Sebastian and Kiela, Douwe},
  editor    = {Larochelle, H and Ranzato, M and Hadsell, R and Balcan, M F and
               Lin, H},
  booktitle = {Advances in Neural Information Processing Systems},
  publisher = {Curran Associates, Inc.},
  volume    = 33,
  pages     = {9459--9474},
  year      = 2020
}

@article{Dhingra2022-in,
  title     = {Time-Aware Language Models as Temporal Knowledge Bases},
  author    = {Dhingra, Bhuwan and Cole, Jeremy R and Eisenschlos, Julian Martin
               and Gillick, Daniel and Eisenstein, Jacob and Cohen, William W},
  editor    = {Roark, Brian and Nenkova, Ani},
  journal   = {Transactions of the Association for Computational Linguistics},
  publisher = {MIT Press},
  address   = {Cambridge, MA},
  volume    = 10,
  pages     = {257--273},
  year      = 2022
}

@inproceedings{Hartvigsen2023-wj,
  title     = {Aging with {GRACE}: Lifelong Model Editing with Discrete
               Key-Value Adaptors},
  author    = {Hartvigsen, Tom and Sankaranarayanan, Swami and Palangi, Hamid
               and Kim, Yoon and Ghassemi, Marzyeh},
  editor    = {Oh, A and Naumann, T and Globerson, A and Saenko, K and Hardt, M
               and Levine, S},
  booktitle = {Advances in Neural Information Processing Systems},
  publisher = {Curran Associates, Inc.},
  volume    = 36,
  pages     = {47934--47959},
  year      = 2023
}

@article{Li2024-hc,
  title     = {{PMET}: Precise model editing in a transformer},
  author    = {Li, Xiaopeng and Li, Shasha and Song, Shezheng and Yang, Jing and
               Ma, Jun and Yu, Jie},
  journal   = {Proc. Conf. AAAI Artif. Intell.},
  publisher = {Association for the Advancement of Artificial Intelligence (AAAI)},
  volume    = 38,
  number    = 17,
  pages     = {18564--18572},
  month     = mar,
  year      = 2024
}

@article{kamath2025tracing,
  author  = {Kamath, Harish and Ameisen, Emmanuel and Kauvar, Isaac and Luger, Rodrigo and Gurnee, Wes and Pearce, Adam and Zimmerman, Sam and Batson, Joshua and Conerly, Thomas and Olah, Chris and Lindsey, Jack},
  title   = {Tracing Attention Computation Through Feature Interactions},
  journal = {Transformer Circuits Thread},
  year    = {2025},
  url     = {https://transformer-circuits.pub/2025/attention-qk/index.html}
}

\appendix


\section{Datasets}
\subsection{ChronosAlign Dataset}
\label{sec:appendix_chronos_alignment_dataset}

Using the ChronosAlign dataset we create three types of questions. (1) Invariant, doesn't specify any temporal condition. (2) Relative, suggests the most recent. (3) Explicit, provides a year to draw information from.
\vspace{2ex}

\begin{itemize}
  \item \textit{Invariant question:} Who won the MTV Europe Music Award for Best Video? Answer:
  \item \textit{Relative question:} Who recently won the MTV Europe Music Award for Best Video? Answer:
  \item \textit{Explicit question:} In 2024, who won the MTV Europe Music Award for Best Video? Answer:
\end{itemize}

\vspace{5ex}

\begin{figure}[hbt!]
  \centering
  \includegraphics[width=0.8\linewidth]{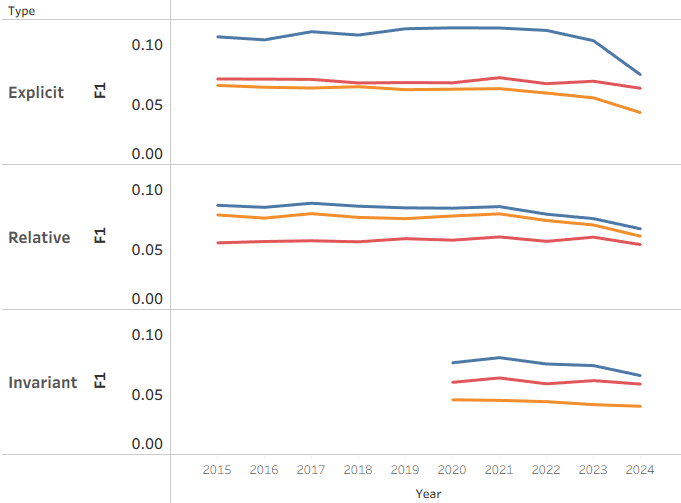}
  \caption{F1 scores for ChronosAlign. We run explicit and relative scores. Gemma 2 (Blue), Red Qwen 3 (Red), LLaMA 3.2B (Orange).}
  \label{fig:chronosalign}
\end{figure}

F1 scores across the entire dataset for 2015 and 2024 are relatively consistent across all models. As each model answers questions for 2023 and 2024 we notice a dip in F1 scores across all three question types. This is expected and is most likely due to training data cut off for each model. However, the most surprising finding from these data is the lower F1 score for Qwen3-4B across all question variants.

For our experiments involving ChronosAlign, we use a random 100 high confidence question answer pairs that have a unique question and are not part of the circuit creation. This increases the sensitivity of F1 and logit difference scores.

\subsection{Taqa dataset}
\label{app:taqa_dataset}

To provide an independent set of question-answer pairs over a wider period of time we incorporate Taqa \citep{Zhao2024-li}; selecting 200 random questions from 2004, 2015, and 2020--2024. We select these years to provide a comparison to ChronosAlign, but also as an independent out of period sample to test our ablations and steering against.

\section{Transcoder Circuit Tracing}

\subsection{Graph Parameters}
\label{sec:appendix_graph_parameters}

Below we outline the parameters used in transcoder circuit tracing.
For extremely high and high sampling confidence rates we used the below cutoff values ($T$).

  \begin{table}[!hbt]
    \centering
    \begin{tabular}{lccc}
      \hline
      & \textbf{Gemma} & \textbf{LLaMA} & \textbf{Qwen} \\
      \hline
      \textbf{} & 0.9 & 0.7 & 0.9 \\
      \hline
    \end{tabular}
    \caption{Extremely high confidence thresholds for each model}
    \label{tab:confidence_thresholds}
  \end{table}

We then sampled up to 20 at random from the extremely high confidence set, for each of the years (2020-2024) to generate graphs.

\subsection{Replacement and Completeness Scores}
\label{appendix:replacementandcompleteness}

\begin{table}[!hbt]
  \centering
  \caption{Average replacement (Repl.) and completeness (Comp.) scores across 2020--2024 for all models. The combined score is the average across all years.}
  \label{tab:replacement_completeness_scores}
  \resizebox{\columnwidth}{!}{%
  \begin{tabular}{lccccccc}
    \hline
    & & \textbf{2020} & \textbf{2021} & \textbf{2022} & \textbf{2023} & \textbf{2024} & \textbf{Combined} \\
    \hline
    \multirow{2}{*}{\textbf{Gemma 2 2B}} & \textbf{Repl.} & 0.732 & 0.730 & 0.730 & 0.729 & 0.733 & 0.731 \\
     & \textbf{Comp.} & 0.929 & 0.928 & 0.929 & 0.928 & 0.929 & 0.929 \\
    \hline
    \multirow{2}{*}{\textbf{LLaMA 3.2 1B}} & \textbf{Repl.} & 0.590 & 0.610 & 0.594 & 0.610 & 0.612 & 0.599 \\
     & \textbf{Comp.} & 0.874 & 0.879 & 0.874 & 0.879 & 0.880 & 0.876 \\
    \hline
    \multirow{2}{*}{\textbf{Qwen3-4b}} & \textbf{Repl.} & 0.462 & 0.461 & 0.462 & 0.458 & 0.464 & 0.461 \\
     & \textbf{Comp.} & 0.884 & 0.884 & 0.884 & 0.883 & 0.885 & 0.884 \\
    \hline
  \end{tabular}%
  }
\end{table}

\subsection{Ablation and steering parameters}
\label{appendix:ablation and steering parameters}

\begin{table}[h]
\centering

\begin{tabular}{lccc|ccc}
\hline
\textbf{Model} & \multicolumn{3}{c|}{\textbf{Ablation}} & \multicolumn{3}{c}{\textbf{Steering}} \\
\cline{2-7}
& \textbf{a} & \textbf{m} & \textbf{l} & \textbf{a} & \textbf{m} & \textbf{l} \\
\hline
Gemma 2 2B   & 1   & -2  & 4 & 1   & 5 & 4 \\
LLaMA 3.2 1B & 0   & -1  & 1 & 0.5 & 5 & 2 \\
Qwen3-4b     & 0.5 & -1  & 4 & 1   & 2 & 4 \\
\hline
\end{tabular}

\caption{parameters used for causal intervention. From Eq. \ref{eq:steering} where $a$ and $m$ are the additive and multiplier coefficients, and $l$ denotes the initial layer intervention unless otherwise specified. The intervention was applied to all tokens except the last two which are the answer block ``Answer:''.}
\label{tab:intervention_parameters}
\end{table}

\clearpage

\subsection{Gemma 2 2B Results}
\begin{table}[ht!]
  \centering
\resizebox{\columnwidth}{!}{%
    \begin{tabular}{l|l|ccccc}
      \hline
       & \textbf{Metric} & \textbf{2020} & \textbf{2021} & \textbf{2022} & \textbf{2023} & \textbf{2024} \\
      \hline
      \multirow{4}{*}{F1} & Ablation F1 & 0.324 & 0.373 & 0.298 & 0.359 & 0.441 \\
      & Relative F1 & 0.514 & 0.477 & 0.425 & 0.487 & 0.549 \\
      & Explicit F1 & 0.908 & 0.914 & 0.917 & 0.925 & 0.903 \\
      & Invariant F1 & 0.446 & 0.509 & 0.390 & 0.444 & 0.635 \\
      \hline
      \multirow{3}{*}{Logit Diff} & Explicit Logit Diff & -0.151 & -0.129 & -0.140 & -0.137 & -0.051 \\
      & Relative Logit Diff & -0.064 & -0.014 & -0.044 & -0.023 & -0.027 \\
      & Invariant Logit Diff & 0.001 & 0.001 & 0.001 & 0.001 & 0.000 \\
      \hline
      \multirow{4}{*}{Probs. Change} & Prior ive & -0.241 & -0.218 & -0.197 & -0.224 & -0.175 \\
      & Prior rve & -0.052 & -0.062 & -0.051 & -0.072 & -0.012 \\
      & Ablated ive & -0.089 & -0.089 & -0.056 & -0.085 & -0.124 \\
      & Ablated rve & 0.035 & 0.053 & 0.045 & 0.042 & 0.012 \\
      \hline
      \end{tabular}%
}
      \caption{F1 scores, with logit probability difference for invariant and relative answers. Gemma 2 2B, ChronosAlign ablation using \commontemporal{} nodes}
  \label{tab:table_chqa_100_scores_gemma2}
\end{table}

\begin{table}[ht!]
  \centering
\resizebox{\columnwidth}{!}{%
    \begin{tabular}{l|l|ccccc}
      \hline
       & \textbf{Metric} & \textbf{2020} & \textbf{2021} & \textbf{2022} & \textbf{2023} & \textbf{2024} \\
      \hline
      \multirow{4}{*}{F1} & Ablation F1 & 0.745 & 0.799 & 0.791 & 0.778 & 0.698 \\
      & Relative F1 & 0.514 & 0.477 & 0.425 & 0.487 & 0.549 \\
      & Explicit F1 & 0.908 & 0.914 & 0.917 & 0.925 & 0.903 \\
      & Invariant F1 & 0.446 & 0.509 & 0.390 & 0.444 & 0.635 \\
      \hline
      \multirow{3}{*}{Logit Diff} & Explicit Logit Diff & -0.005 & 0.006 & 0.003 & -0.002 & -0.005 \\
      & Relative Logit Diff & 0.000 & 0.009 & 0.007 & 0.011 & 0.001 \\
      & Invariant Logit Diff & 0.003 & 0.008 & 0.003 & 0.007 & 0.002 \\
      \hline
      \multirow{4}{*}{Probs. Change} & Prior ive & -0.079 & -0.073 & -0.076 & -0.091 & -0.017 \\
      & Prior rve & -0.052 & -0.062 & -0.051 & -0.072 & -0.012 \\
      & Ablated ive & -0.071 & -0.071 & -0.076 & -0.082 & -0.009 \\
      & Ablated rve & -0.048 & -0.059 & -0.047 & -0.059 & -0.006 \\
      \hline
      \end{tabular}%
}
      \caption{F1 scores, with logit probability difference for invariant and relative answers. Gemma 2 2B, ChronosAlign ablation using time invariant nodes.}
  \label{tab:table_chqa_100_scores_random_invariant_gemma2}
\end{table}

\begin{table}[ht!]
  \centering
\resizebox{\columnwidth}{!}{%
    \begin{tabular}{l|l|ccccc}
      \hline
       & \textbf{Metric} & \textbf{2020} & \textbf{2021} & \textbf{2022} & \textbf{2023} & \textbf{2024} \\
      \hline
      \multirow{4}{*}{F1} & Ablation F1 & 0.231 & 0.149 & 0.163 & 0.201 & 0.158 \\
      & Relative F1 & 0.071 & 0.085 & 0.020 & 0.046 & 0.081 \\
      & Explicit F1 & 0.351 & 0.370 & 0.416 & 0.320 & 0.148 \\
      & Invariant F1 & 0.083 & 0.076 & 0.056 & 0.089 & 0.118 \\
      \hline
      \multirow{4}{*}{Logit Diff} & Explicit Logit Diff & 0.079 & 0.051 & 0.039 & 0.086 & 0.019 \\
      & Relative Logit Diff & -0.042 & -0.039 & -0.031 & -0.060 & -0.066 \\
      & Invariant Logit Diff & -0.030 & -0.028 & -0.023 & -0.049 & -0.047 \\
      & AE Logit Diff & 0.116 & 0.071 & 0.054 & 0.101 & 0.119 \\
      \hline
      \multirow{4}{*}{Probs. Change} & Prior ive & 0.045 & 0.055 & 0.051 & 0.049 & 0.022 \\
      & Prior rve & 0.063 & 0.073 & 0.069 & 0.068 & 0.041 \\
      & Steered ive & -0.064 & -0.024 & -0.011 & -0.086 & -0.043 \\
      & Steered rve & -0.058 & -0.016 & -0.000 & -0.079 & -0.044 \\
      \hline
      \end{tabular}%
}
      \caption{F1 scores, with logit probability difference for invariant and relative answers. Gemma 2 2B, ChronosAlign Steering using \commontoyear{} nodes}
  \label{tab:table_chqa_100_scaled_steering _gemma2}
\end{table}

\begin{table}[hbt!]
  \centering
  \resizebox{\columnwidth}{!}{%
  \begin{tabular}{l|l|ccccc}
    \hline
     & \textbf{Metric} & \textbf{2004} & \textbf{2015} & \textbf{2020} & \textbf{2021} & \textbf{2022} \\
    \hline
    \multirow{4}{*}{F1} & Ablation F1 & 0.168 & 0.226 & 0.157 & 0.159 & 0.140 \\
    & Relative F1 & 0.124 & 0.232 & 0.153 & 0.174 & 0.173 \\
    & Explicit F1 & 0.199 & 0.260 & 0.207 & 0.223 & 0.231 \\
    \hline
    \multirow{2}{*}{Logit Diff} & Explicit Logit Diff & -0.054 & -0.066 & -0.100 & -0.119 & -0.128 \\
    & Relative Logit Diff & -0.008 & -0.035 & -0.027 & -0.016 & -0.015 \\
    \hline
    \multirow{2}{*}{Probs. Change} & Prior rve & -0.082 & -0.032 & -0.039 & -0.056 & -0.065 \\
    & New rve & -0.036 & -0.000 & 0.034 & 0.047 & 0.048 \\
    \hline
  \end{tabular}%
  }
        \caption{F1 scores, with logit probability difference for invariant and relative answers. Gemma 2 2B, Taqa ablation using \commontemporal{} nodes}
  \label{tab:table_taqa_200_gemma_2}
\end{table}

\begin{table}[H]
  \resizebox{\columnwidth}{!}{%

  \begin{tabular}{lccc}
    \textbf{Dataset} & \textbf{Ablation F1} & \textbf{Original F1} & \textbf{Original Answer Logit Diff} \\
      \hline
        Commonsense & 0.377 & 0.370 & 0.007 \\
        GSM8K & 0.307 & 0.262 & -0.002 \\
    \hline
  \end{tabular}%
  }
  \caption{F1 Score and logit probabilities for ablation results on time-invariant datasets, Gemma 2 2B}
  \label{tab:ablation_temporal_invariant_datasets_gemma_2}
\end{table}

\begin{table}[H]
  \centering
  \resizebox{\columnwidth}{!}{%
  \begin{tabular}{l|l|ccccc}
      \hline
       & \textbf{Metric} & \textbf{2020} & \textbf{2021} & \textbf{2022} & \textbf{2023} & \textbf{2024} \\
      \hline
      \multirow{4}{*}{F1} & Ablation F1 & 0.442 & 0.571 & 0.525 & 0.467 & 0.682 \\
      & Relative F1 & 0.514 & 0.477 & 0.425 & 0.487 & 0.549 \\
      & Explicit F1 & 0.908 & 0.914 & 0.917 & 0.925 & 0.903 \\
      & Invariant F1 & 0.446 & 0.509 & 0.390 & 0.444 & 0.635 \\
      \hline
      \multirow{4}{*}{Logit Diff} & Explicit Logit Diff & -0.095 & -0.076 & -0.075 & -0.096 & -0.014 \\
      & Relative Logit Diff & -0.015 & 0.012 & -0.010 & 0.014 & -0.004 \\
      & Invariant Logit Diff & 0.004 & 0.018 & -0.008 & 0.023 & -0.001 \\
      & AE Logit Diff & 0.023 & 0.012 & 0.023 & 0.066 & -0.007 \\
      \hline
      \multirow{4}{*}{Probs. Change} & Prior ive & -0.079 & -0.071 & -0.076 & -0.090 & -0.016 \\
      & Prior rve & -0.052 & -0.060 & -0.052 & -0.072 & -0.011 \\
      & Ablated ive & 0.021 & 0.023 & -0.009 & 0.029 & -0.003 \\
      & Ablated rve & 0.028 & 0.028 & 0.013 & 0.038 & -0.001 \\
      \hline
      \end{tabular}%
  }

      \caption{F1 scores, with logit probability difference for invariant and relative answers. Ablation using higher temporal representation, layers equal to or greater than 12. Gemma 2 2B, ChronosAlign}
  \label{tab:table_chqa_100_high_rep_ablation_scores_gemma2}
\end{table}

\begin{table}[H]
  \centering
  \resizebox{\columnwidth}{!}{%
  \begin{tabular}{l|l|ccccc}
      \hline
       & \textbf{Metric} & \textbf{2020} & \textbf{2021} & \textbf{2022} & \textbf{2023} & \textbf{2024} \\
      \hline
      \multirow{4}{*}{F1} & Ablation F1 & 0.321 & 0.377 & 0.298 & 0.321 & 0.440 \\
      & Relative F1 & 0.514 & 0.477 & 0.425 & 0.487 & 0.549 \\
      & Explicit F1 & 0.908 & 0.914 & 0.917 & 0.925 & 0.903 \\
      & Invariant F1 & 0.446 & 0.509 & 0.390 & 0.444 & 0.635 \\
      \hline
      \multirow{4}{*}{Logit Diff} & Explicit Logit Diff & -0.148 & -0.119 & -0.122 & -0.127 & -0.033 \\
      & Relative Logit Diff & -0.067 & -0.022 & -0.042 & -0.033 & -0.019 \\
      & Invariant Logit Diff & -0.083 & -0.096 & -0.102 & -0.084 & -0.044 \\
      & AE Logit Diff & -0.002 & 0.020 & 0.015 & 0.025 & 0.005 \\
      \hline
      \multirow{4}{*}{Probs. Change} & Prior ive & -0.070 & -0.062 & -0.068 & -0.075 & -0.012 \\
      & Prior rve & -0.054 & -0.062 & -0.049 & -0.057 & -0.012 \\
      & Ablated ive & 0.033 & 0.034 & 0.036 & 0.042 & 0.007 \\
      & Ablated rve & 0.027 & 0.035 & 0.031 & 0.038 & 0.002 \\
      \hline
      \end{tabular}%
  }
      \caption{Semantically different questions, where time is presented at the end of the question. F1 scores, with logit probability difference for invariant and relative answers.}
  \label{tab:table_chqa_semantically_different_gemma2}
\end{table}

\clearpage

\subsection{Qwen3-4b}
\label{app:ablation_Qwen3-4b}

\begin{table}[h]

  \resizebox{\columnwidth}{!}{%
    \begin{tabular}{l|l|ccccc}
      \hline
       & \textbf{Metric} & \textbf{2020} & \textbf{2021} & \textbf{2022} & \textbf{2023} & \textbf{2024} \\
      \hline
      \multirow{4}{*}{F1} & Ablation F1 & 0.404 & 0.491 & 0.395 & 0.437 & 0.409 \\
      & Relative F1 & 0.557 & 0.549 & 0.539 & 0.617 & 0.460 \\
      & Explicit F1 & 0.896 & 0.897 & 0.887 & 0.907 & 0.911 \\
      & Invariant F1 & 0.577 & 0.587 & 0.554 & 0.628 & 0.507 \\
      \hline
      \multirow{3}{*}{Logit Diff} & Explicit Logit Diff & -0.245 & -0.260 & -0.238 & -0.211 & -0.194 \\
      & Relative Logit Diff & -0.140 & -0.119 & -0.136 & -0.087 & -0.077 \\
      & Invariant Logit Diff & -0.083 & -0.096 & -0.102 & -0.084 & -0.044 \\
      \hline
      \multirow{4}{*}{Probs. Change} & Prior ive & -0.124 & -0.129 & -0.098 & -0.075 & -0.112 \\
      & Prior rve & -0.120 & -0.158 & -0.133 & -0.097 & -0.166 \\
      & Ablated ive & 0.037 & 0.035 & 0.038 & 0.053 & 0.038 \\
      & Ablated rve & -0.015 & -0.018 & -0.031 & 0.027 & -0.049 \\
      \hline
      \end{tabular}%
  }
      \caption{F1 scores, with logit probability difference for invariant and relative answers. Qwen3-4b, ChronosAlign ablation using \commontemporal{} nodes}
  \label{tab:table_chqa_100_scores_qwen3}
\end{table}

\begin{table}[h!]
    \resizebox{\columnwidth}{!}{%
    \begin{tabular}{l|l|ccccc}
      \hline
       & \textbf{Metric} & \textbf{2020} & \textbf{2021} & \textbf{2022} & \textbf{2023} & \textbf{2024} \\
      \hline
      \multirow{4}{*}{F1} & Ablation F1 & 0.475 & 0.538 & 0.383 & 0.490 & 0.421 \\
      & Relative F1 & 0.557 & 0.549 & 0.539 & 0.617 & 0.460 \\
      & Explicit F1 & 0.896 & 0.897 & 0.887 & 0.907 & 0.911 \\
      & Invariant F1 & 0.577 & 0.587 & 0.554 & 0.628 & 0.507 \\
      \hline
      \multirow{3}{*}{Logit Diff} & Explicit Logit Diff & -0.189 & -0.207 & -0.228 & -0.190 & -0.218 \\
      & Relative Logit Diff & -0.105 & -0.130 & -0.123 & -0.068 & -0.081 \\
      & Invariant Logit Diff & -0.114 & -0.131 & -0.138 & -0.134 & -0.121 \\
      \hline
      \multirow{4}{*}{Probs. Change} & Prior ive & -0.124 & -0.129 & -0.098 & -0.075 & -0.112 \\
      & Prior rve & -0.120 & -0.158 & -0.133 & -0.097 & -0.166 \\
      & Ablated ive & -0.050 & -0.053 & -0.009 & -0.019 & -0.015 \\
      & Ablated rve & -0.036 & -0.082 & -0.029 & 0.025 & -0.028 \\
      \hline
    \end{tabular}%
    }
    \caption{F1 and cumulative probability scores for 2020-2024 given the top 100 (Qwen3-4b, ChronosAlign random invariant)}
  \label{tab:table_chqa_100_scores_qwen3_invariant}
\end{table}

\begin{table}[h!]
  \resizebox{\columnwidth}{!}{%
    \begin{tabular}{l|l|ccccc}
      \hline
       & \textbf{Metric} & \textbf{2020} & \textbf{2021} & \textbf{2022} & \textbf{2023} & \textbf{2024} \\
      \hline
      \multirow{4}{*}{F1} & Ablation F1 & 0.099 & 0.099 & 0.072 & 0.074 & 0.104 \\
      & Relative F1 & 0.086 & 0.073 & 0.079 & 0.085 & 0.089 \\
      & Explicit F1 & 0.256 & 0.325 & 0.241 & 0.268 & 0.288 \\
      & Invariant F1 & 0.113 & 0.113 & 0.128 & 0.098 & 0.152 \\
      \hline
      \multirow{3}{*}{Logit Diff} & Explicit Logit Diff & 0.011 & 0.013 & 0.000 & 0.002 & 0.002 \\
      & Relative Logit Diff & 0.000 & 0.004 & 0.000 & 0.002 & -0.003 \\
      & Invariant Logit Diff & 0.009 & 0.013 & 0.000 & 0.003 & 0.006 \\
      \hline
      \multirow{4}{*}{Probs. Change} & Prior ive & 0.122 & 0.077 & 0.061 & 0.047 & 0.064 \\
      & Prior rve & 0.338 & 0.303 & 0.280 & 0.273 & 0.286 \\
      & Ablated ive & 0.121 & 0.077 & 0.060 & 0.048 & 0.068 \\
      & Ablated rve & 0.327 & 0.294 & 0.280 & 0.273 & 0.281 \\
      \hline
      \end{tabular}%
  }
      \caption{F1 scores, with logit probability difference for invariant and relative answers. Qwen3-4b, ChronosAlign steering using \commontoyear{} nodes}
  \label{tab:table_chqa_100_scores_steering_qwen3}
\end{table}

\begin{table}[h!]
  \resizebox{\columnwidth}{!}{%
  \begin{tabular}{l|l|ccccc}
    \hline
     & \textbf{Metric} & \textbf{2004} & \textbf{2015} & \textbf{2020} & \textbf{2021} & \textbf{2022} \\
    \hline
    \multirow{4}{*}{F1} & Ablation F1 & 0.168 & 0.226 & 0.157 & 0.159 & 0.140 \\
    & Relative F1 & 0.124 & 0.232 & 0.153 & 0.174 & 0.173 \\
    & Explicit F1 & 0.199 & 0.260 & 0.207 & 0.223 & 0.231 \\
    \hline
    \multirow{2}{*}{Logit Diff} & Explicit Logit Diff & -0.054 & -0.066 & -0.100 & -0.119 & -0.128 \\
    & Relative Logit Diff & -0.008 & -0.035 & -0.027 & -0.016 & -0.015 \\
    \hline
    \multirow{2}{*}{Probs. Change} & Prior rve & -0.082 & -0.032 & -0.039 & -0.056 & -0.065 \\
    & New rve & -0.036 & -0.000 & 0.034 & 0.047 & 0.048 \\
    \hline
  \end{tabular}%
  }
  \caption{F1 scores, with logit probability difference for invariant and relative answers. Qwen3-4b, Taqa ablation using \commontemporal{} nodes}
  \label{tab:table_taqa_200_qwen3}
\end{table}

\begin{table}[h]
    \resizebox{\columnwidth}{!}{%
  \begin{tabular}{lccc}
    \textbf{Dataset} & \textbf{Ablation F1} & \textbf{Original F1} & \textbf{Original Answer Logit Diff} \\
      \hline
        Commonsense & 0.457 & 0.572 & -0.087 \\
        GSM8K & 0.310 & 0.515 & -0.024 \\
    \hline
  \end{tabular}%
    }
  \caption{F1 Score and logit probabilities for ablation results on time-invariant datasets, Qwen3-4b}
  \label{tab:ablation_temporal_invariant_nodes_datasets_qwen3}
\end{table}

\begin{table}[h]
    \resizebox{\columnwidth}{!}{%
  \begin{tabular}{l|l|ccccc}
      \hline
       & \textbf{Metric} & \textbf{2020} & \textbf{2021} & \textbf{2022} & \textbf{2023} & \textbf{2024} \\
      \hline
      \multirow{4}{*}{F1} & Ablation F1 & 0.573 & 0.755 & 0.558 & 0.683 & 0.554 \\
      & Relative F1 & 0.557 & 0.549 & 0.539 & 0.617 & 0.460 \\
      & Explicit F1 & 0.896 & 0.897 & 0.887 & 0.907 & 0.911 \\
      & Invariant F1 & 0.577 & 0.587 & 0.554 & 0.628 & 0.507 \\
      \hline
      \multirow{4}{*}{Logit Diff} & Explicit Logit Diff & -0.108 & -0.091 & -0.084 & -0.057 & -0.140 \\
      & Relative Logit Diff & -0.051 & -0.043 & -0.019 & -0.003 & -0.023 \\
      & Invariant Logit Diff & -0.057 & -0.044 & -0.034 & -0.045 & -0.051 \\
      & AE Logit Diff & 0.063 & 0.039 & 0.070 & 0.086 & 0.112 \\
      \hline
      \multirow{4}{*}{Probs. Change} & Prior ive & -0.124 & -0.129 & -0.098 & -0.075 & -0.112 \\
      & Prior rve & -0.120 & -0.158 & -0.133 & -0.097 & -0.166 \\
      & Ablated ive & -0.072 & -0.082 & -0.048 & -0.063 & -0.023 \\
      & Ablated rve & -0.062 & -0.111 & -0.069 & -0.043 & -0.049 \\
      \hline
      \end{tabular}%
    }
      \caption{F1 scores, with logit probability difference for invariant and relative answers. Ablation using higher temporal representations, \texttt{L11F148703}, \texttt{L15F127839}, \texttt{L16F45664}. Qwen3 4b, ChronosAlign}
  \label{tab:table_chqa_100_high_rep_ablation_scores_qwen3}
\end{table}

\begin{table}[h]
  \resizebox{\columnwidth}{!}{%
  \begin{tabular}{l|l|ccccc}
      \hline
       & \textbf{Metric} & \textbf{2020} & \textbf{2021} & \textbf{2022} & \textbf{2023} & \textbf{2024} \\
      \hline
      \multirow{4}{*}{F1} & Ablation F1 & 0.356 & 0.402 & 0.324 & 0.493 & 0.359 \\
      & Relative F1 & 0.557 & 0.549 & 0.539 & 0.617 & 0.460 \\
      & Explicit F1 & 0.896 & 0.897 & 0.887 & 0.907 & 0.911 \\
      & Invariant F1 & 0.577 & 0.587 & 0.554 & 0.628 & 0.507 \\
      \hline
      \multirow{4}{*}{Logit Diff} & Explicit Logit Diff & -0.266 & -0.260 & -0.255 & -0.165 & -0.278 \\
      & Relative Logit Diff & -0.096 & -0.086 & -0.125 & -0.011 & -0.124 \\
      & Invariant Logit Diff & -0.069 & -0.058 & -0.099 & 0.004 & -0.077 \\
      & AE Logit Diff & 0.172 & 0.156 & 0.230 & 0.176 & 0.160 \\
      \hline
      \multirow{4}{*}{Probs. Change} & Prior ive & -0.147 & -0.146 & -0.126 & -0.128 & -0.193 \\
      & Prior rve & -0.185 & -0.176 & -0.169 & -0.157 & -0.196 \\
      & Ablated ive & 0.051 & 0.055 & 0.030 & 0.041 & 0.009 \\
      & Ablated rve & -0.014 & -0.002 & -0.039 & -0.003 & -0.041 \\
      \hline
      \end{tabular}%
  }
      \caption{Semantically different questions, where time is presented at the end of the question. F1 scores, with logit probability difference for invariant and relative answers.}
  \label{tab:table_chqa_semantically_different_qwen3}
\end{table}

\begin{figure}[h]
  \centering
  \includegraphics[width=0.8\linewidth]{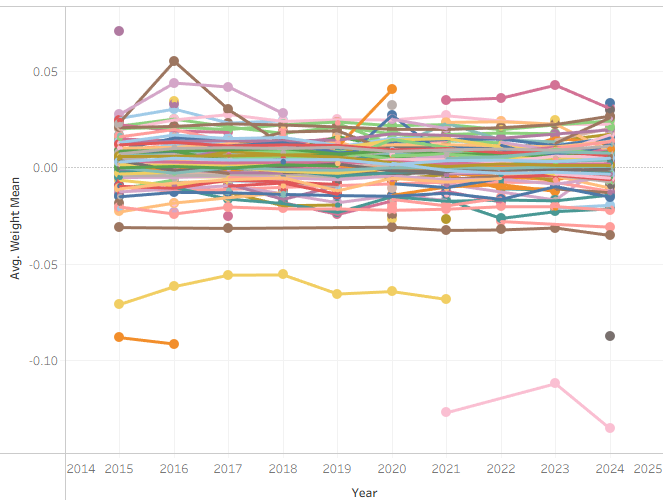}
  \caption{Qwen3-4b, higher layer nodes. Mean edge weights (y-axis), by years (x-axis) highlighting temporal anchors across a 2015-2024 period. Demonstrates how a variety of nodes regarding time have a larger (or lower) edge weight compared to other temporal nodes. Specifically L11F148703 (yellow), L10F125625 (orange), L16F45664 (pink) and L15F127829 (grey)}.
  \label{fig:TemporalNodes_FeatureMapping_qwen}
\end{figure}

\clearpage

\subsection{LLaMA 3.2 1B}
\label{app:ablation_LLaMA 3.2 1B}

\begin{table}[h]
  \resizebox{\columnwidth}{!}{%
    \begin{tabular}{l|l|ccccc}
      \hline
       & \textbf{Metric} & \textbf{2020} & \textbf{2021} & \textbf{2022} & \textbf{2023} & \textbf{2024} \\
      \hline
      \multirow{4}{*}{F1} & Ablation F1 & 0.496 & 0.383 & 0.393 & 0.425 & 0.653 \\
      & Relative F1 & 0.584 & 0.560 & 0.469 & 0.525 & 0.493 \\
      & Explicit F1 & 0.988 & 0.967 & 0.969 & 0.981 & 0.965 \\
      & Invariant F1 & 0.464 & 0.408 & 0.365 & 0.393 & 0.599 \\
      \hline
      \multirow{4}{*}{Logit Diff} & Explicit Logit Diff & -0.129 & -0.146 & -0.128 & -0.114 & -0.026 \\
      & Relative Logit Diff & -0.082 & -0.078 & -0.039 & -0.038 & -0.008 \\
      & Invariant Logit Diff & 0.003 & 0.017 & -0.001 & 0.027 & -0.005 \\
      & AE Logit Diff & -0.003 & 0.009 & 0.002 & 0.024 & 0.006 \\
      \hline
      \multirow{4}{*}{Probs. Change} & Prior ive & -0.106 & -0.124 & -0.092 & -0.103 & -0.020 \\
      & New ive & 0.027 & 0.039 & 0.036 & 0.038 & 0.000 \\
      & Prior rve & -0.051 & -0.056 & -0.087 & -0.072 & -0.049 \\
      & New rve & -0.004 & 0.011 & 0.002 & 0.004 & -0.031 \\
      \hline
      \end{tabular}%
  }
      \caption{F1 scores, with logit probability difference for invariant and relative answers. LLaMA 3.2 1B, ChronosAlign ablation using \commontemporal{} nodes}
  \label{tab:table_chqa_100_scores_llama3}
\end{table}

\begin{table}[ht!]
  \resizebox{\columnwidth}{!}{%
    \begin{tabular}{l|l|ccccc}
      \hline
       & \textbf{Metric} & \textbf{2020} & \textbf{2021} & \textbf{2022} & \textbf{2023} & \textbf{2024} \\
      \hline
      \multirow{4}{*}{F1} & Ablation F1 & 0.003 & 0.003 & 0.002 & 0.005 & 0.014 \\
      & Relative F1 & 0.584 & 0.560 & 0.469 & 0.525 & 0.493 \\
      & Explicit F1 & 0.988 & 0.967 & 0.969 & 0.981 & 0.965 \\
      & Invariant F1 & 0.464 & 0.408 & 0.365 & 0.393 & 0.599 \\
      \hline
      \multirow{4}{*}{Logit Diff} & Explicit Logit Diff & -0.318 & -0.301 & -0.236 & -0.259 & -0.177 \\
      & Relative Logit Diff & -0.264 & -0.245 & -0.144 & -0.182 & -0.129 \\
      & Invariant Logit Diff & -0.211 & -0.176 & -0.140 & -0.155 & -0.154 \\
      & AE Logit Diff & 0.112 & 0.113 & 0.110 & 0.108 & 0.064 \\
      \hline
      \multirow{4}{*}{Probs. Change} & Prior ive & -0.106 & -0.124 & -0.092 & -0.103 & -0.020 \\
      & New ive & 0.002 & 0.001 & 0.004 & 0.002 & 0.002 \\
      & Prior rve & -0.051 & -0.056 & -0.087 & -0.072 & -0.049 \\
      & New rve & 0.003 & -0.000 & 0.005 & 0.006 & -0.002 \\
      \hline
    \end{tabular}%
  }
    \caption{F1 and cumulative probability scores for 2020-2024 given the top 100 (LLaMA 3.2 1B, ChronosAlign random invariant)}
  \label{tab:table_chqa_100_scores_llama32_invariant}
\end{table}

\begin{table}[ht!]
    \resizebox{\columnwidth}{!}{%

    \begin{tabular}{l|l|ccccc}
      \hline
       & \textbf{Metric} & \textbf{2020} & \textbf{2021} & \textbf{2022} & \textbf{2023} & \textbf{2024} \\
      \hline
      \multirow{4}{*}{F1} & Steer F1 & 0.089 & 0.136 & 0.154 & 0.109 & 0.033 \\
      & Relative F1 & 0.126 & 0.149 & 0.158 & 0.172 & 0.108 \\
      & Explicit F1 & 0.252 & 0.400 & 0.425 & 0.204 & 0.070 \\
      & Invariant F1 & 0.053 & 0.069 & 0.107 & 0.058 & 0.076 \\
      \hline
      \multirow{4}{*}{Logit Diff} & Explicit Logit Diff & 0.030 & 0.000 & 0.023 & 0.012 & -0.069 \\
      & Relative Logit Diff & -0.025 & -0.020 & 0.001 & -0.011 & -0.047 \\
      & Invariant Logit Diff & -0.056 & -0.046 & -0.033 & -0.038 & -0.072 \\
      & AE Logit Diff & 0.075 & 0.048 & 0.043 & 0.052 & 0.007 \\
      \hline
      \multirow{4}{*}{Probs. Change} & Prior ive & 0.032 & 0.013 & 0.028 & 0.021 & -0.002 \\
      & New ive & -0.054 & -0.034 & -0.028 & -0.029 & -0.006 \\
      & Prior rve & 0.007 & -0.012 & 0.003 & -0.004 & -0.027 \\
      & New rve & -0.048 & -0.032 & -0.019 & -0.027 & -0.005 \\
      \hline
      \end{tabular}%
    }
      \caption{F1 scores, with logit probability difference for invariant and relative answers. LLaMA 3.2 1B, ChronosAlign steering using \commontoyear{} nodes}
  \label{tab:table_chqa_100_scores_steering_llama3}
\end{table}

\begin{table}[h]
    \resizebox{\columnwidth}{!}{%
  \begin{tabular}{l|l|ccccc}
    \hline
     & \textbf{Metric} & \textbf{2004} & \textbf{2015} & \textbf{2020} & \textbf{2021} & \textbf{2022} \\
    \hline
    \multirow{4}{*}{F1} & Ablation F1 & 0.014 & 0.032 & 0.028 & 0.032 & 0.031 \\
    & Relative F1 & 0.016 & 0.029 & 0.025 & 0.032 & 0.021 \\
    & Explicit F1 & 0.022 & 0.032 & 0.027 & 0.048 & 0.036 \\
    \hline
    \multirow{2}{*}{Logit Diff} & Explicit Logit Diff & -0.014 & -0.030 & -0.057 & -0.061 & -0.055 \\
    & Relative Logit Diff & -0.009 & -0.015 & -0.012 & -0.010 & -0.007 \\
    \hline
    \multirow{2}{*}{Probs. Change} & Prior rve & -0.028 & -0.013 & -0.028 & -0.032 & -0.029 \\
    & New rve & -0.023 & 0.001 & 0.017 & 0.020 & 0.018 \\
    \hline
  \end{tabular}%
    }
  \caption{F1 scores, with logit probability difference for invariant and relative answers. LLaMA 3.2 1B, Taqa ablation using \commontemporal{} nodes}
  \label{tab:table_taqa_200_llama3}
\end{table}

\begin{table}[hbt!]
  \resizebox{\columnwidth}{!}{%
  \begin{tabular}{lccc}
    \textbf{Dataset} & \textbf{Ablation F1} & \textbf{Original F1} & \textbf{Original Answer Logit Diff} \\
      \hline
        Commonsense & 0.304 & 0.310 & 0.003 \\
        GSM8K & 0.045 & 0.043 & -0.001 \\
    \hline
  \end{tabular}%
  }
  \caption{F1 Score and logit probabilities for ablation results on time-invariant datasets, LLaMA 3.2 1B}
  \label{tab:ablation_temporal_invariant_nodes_datasets_llama3}
\end{table}

\begin{table}[h]
  \resizebox{\columnwidth}{!}{%
  \begin{tabular}{l|l|ccccc}
      \hline
       & \textbf{Metric} & \textbf{2020} & \textbf{2021} & \textbf{2022} & \textbf{2023} & \textbf{2024} \\
      \hline
      \multirow{4}{*}{F1} & Ablation F1 & 0.497 & 0.332 & 0.337 & 0.347 & 0.569 \\
      & Relative F1 & 0.584 & 0.560 & 0.469 & 0.525 & 0.493 \\
      & Explicit F1 & 0.988 & 0.967 & 0.969 & 0.981 & 0.965 \\
      & Invariant F1 & 0.464 & 0.408 & 0.365 & 0.393 & 0.599 \\
      \hline
      \multirow{4}{*}{Logit Diff} & Explicit Logit Diff & -0.119 & -0.121 & -0.109 & -0.105 & -0.028 \\
      & Relative Logit Diff & -0.073 & -0.050 & -0.037 & -0.033 & -0.013 \\
      & Invariant Logit Diff & -0.000 & 0.024 & 0.005 & 0.016 & 0.004 \\
      & AE Logit Diff & -0.008 & 0.041 & 0.026 & 0.041 & 0.011 \\
      \hline
      \multirow{4}{*}{Probs. Change} & Prior ive & -0.097 & -0.098 & -0.075 & -0.082 & -0.016 \\
      & Prior rve & -0.055 & -0.056 & -0.078 & -0.063 & -0.029 \\
      & Ablated ive & 0.022 & 0.048 & 0.040 & 0.039 & 0.017 \\
      & Ablated rve & -0.008 & 0.016 & -0.006 & 0.009 & -0.013 \\
      \hline
      \end{tabular}%
        }
      \caption{Semantically different questions, where time is presented at the end of the question. F1 scores, with logit probability difference for invariant and relative answers. LLaMA 3.2 1B, ChronosAlign dataset.}
  \label{tab:table_chqa_semantically_different_llama3}
\end{table}

\begin{figure}[hbt!]
  \centering
  \includegraphics[width=0.8\linewidth]{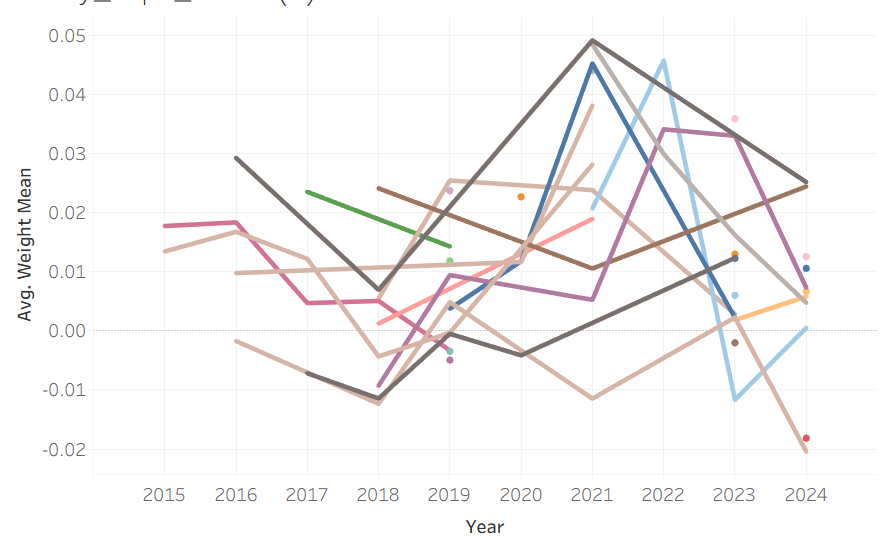}
  \caption{LLaMA 3.2 1B, layers between 4 and 8. Mean edge weights (y-axis), by years (x-axis) highlighting temporal anchors across a temporal continuum. Demonstrates a convoluted representation of time nodes. Disentangling temporal representations for LLaMA 3.2 1B is difficult}
  \label{fig:Lama32_TemporalNodes_FeatureMapping_2015_2024}
\end{figure}

\clearpage

\section{Transcoder Circuit tracing examples}
\label{sec:appendix_circuit_tracing_examples}

We generate graphs which are limited to 17000 nodes, for simple temporal statements. A 5000 node example highlighting \commontemporal{} and \commontoyear{} nodes can be found here: \href{https://www.neuronpedia.org/gemma-2-2b/graph?slug=in2021whoisthepr-1771220421091&pruningThreshold=0.8&densityThreshold=0.99}{Neuronpedia graph}.


\subsection{Token Activation Testing for Higher Temporal Representations} 
\label{app:Activation_Testing_for_Higher_Temporal_Representations}

To test for higher temporal representation features we generated a set of time specific data; prompting Google Gemini 3 for a list of unique and iconic places, people and events from 2015 to 2024. The reference data can be found in the repository: \url{https://github.com/sg-sy/timewarp-temporaltranscoder}. 

We then calculate the average activation across years for each prompt. For each year $y$, layer $l$, and feature $f$, we compute the average activation across all tokens $T_y$ within a single prompt associated with that year:

\begin{equation}
\bar{A}_{y,l,f} = \frac{1}{|T_y|} \sum_{t \in T_y} a_{t,l,f}
\end{equation}

\noindent where $a_{t,l,f}$ denotes the activation value for token $t$ at layer $l$ and feature $f$. We then normalise these average activations across years using min-max normalisation:

\begin{equation}
\hat{A}_{y,l,f} = \frac{\bar{A}_{y,l,f} - \min_{y'} \bar{A}_{y',l,f}}{\max_{y'} \bar{A}_{y',l,f} - \min_{y'} \bar{A}_{y',l,f}}
\end{equation}

\noindent where $y'$ ranges over all years in the evaluation set. This normalisation ensures that activation patterns are comparable across different temporal periods, revealing features that selectively respond to specific years.

For the figures below we filter to layer, feature id combinations that have higher influence scores, or in the case of LLaMA 3.2 1B present temporal token activations that align with the patterns exhibited by Qwen3 and Gemma2. Testing token activations with these texts reveals that \textit{chrono-semantic} nodes indeed capture a range of semantic meaning from that period of time that goes beyond simple numerical parsing.

Gemma 2 2B presents an additive methodology whereby temporal representation is created from adding temporal concepts to the underlying representations. On the other hand Qwen3-4b presents a subtractive methodology, using components to subtract from the underlying representation of time created. For Qwen3-4b, this presents as a high invariant value, with relatively high year values for years it aims to remove from the answer. Layer 16, feature 45664 presents a good example, whereby this node is present for the year 2023 and 2024 due to its ablative properties for tokens that activate for invariant, and years below 2023.

\begin{figure}[h]
  \centering
  \includegraphics[width=0.8\linewidth]{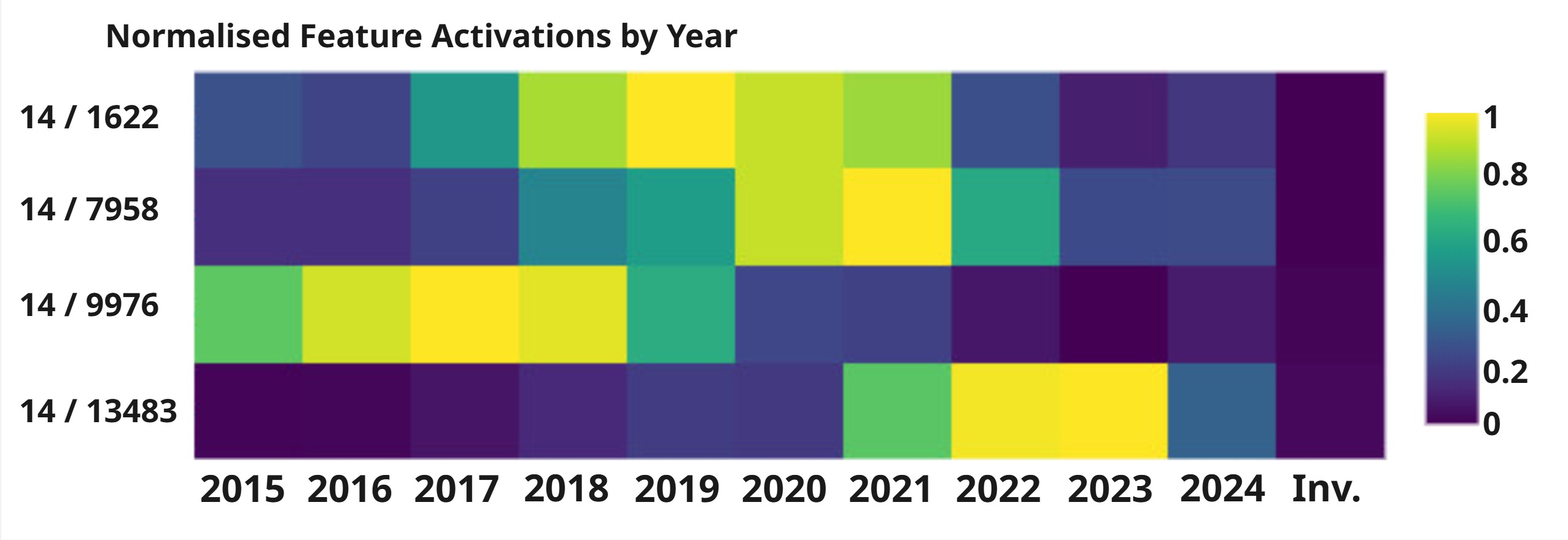}
  \caption{Gemma 2 2B activation heatmap for chrono-semantic specialised features. Note that the highest values for each feature span one to two years with no activation for the invariant data (last column)}
  \label{fig:Gemma_activation_heatmap_chronosemantic_specialised}
\end{figure}

\begin{figure}[h]
  \centering
  \includegraphics[width=0.8\linewidth]{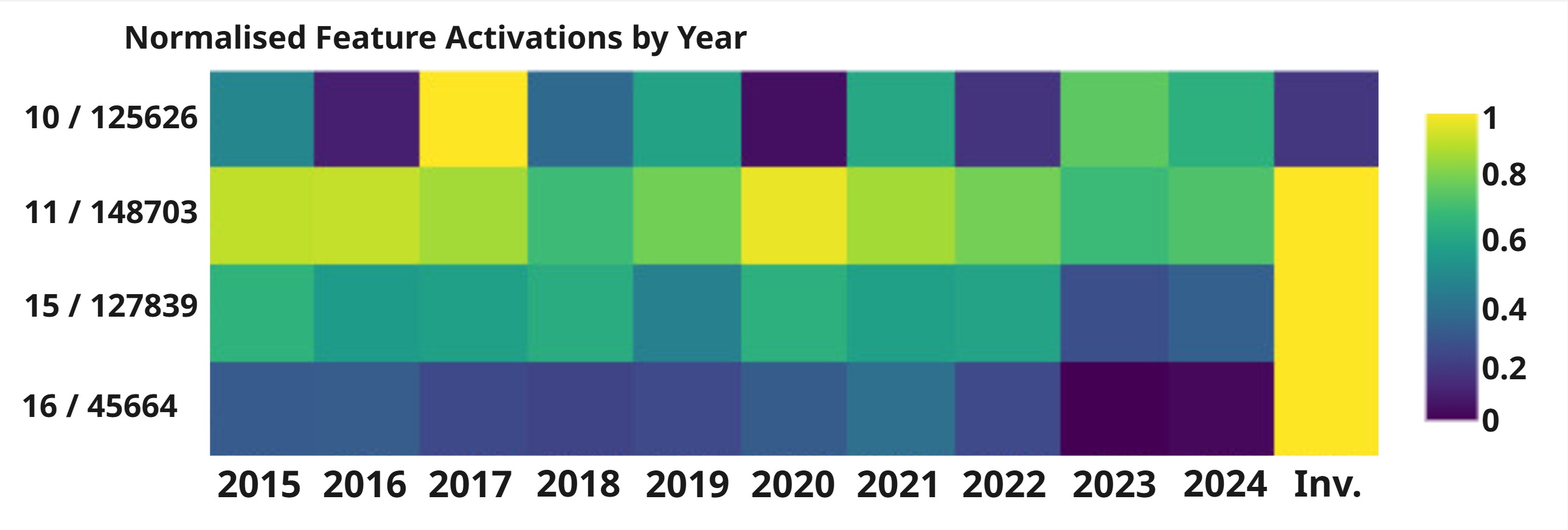}
  \caption{Qwen3-4b activation heatmap for chrono-semantic specialised features. The highest activations occur for the invariant column, with varying activations across yearly representations. This graph is the inverse of Gemma 2 2B, whereby the darker colours highlight which years are to be aligned to.}
  \label{fig:Qwen_activation_heatmap_chronosemantic_specialised}
\end{figure}

\begin{figure}[h]
  \centering
  \includegraphics[width=0.8\linewidth]{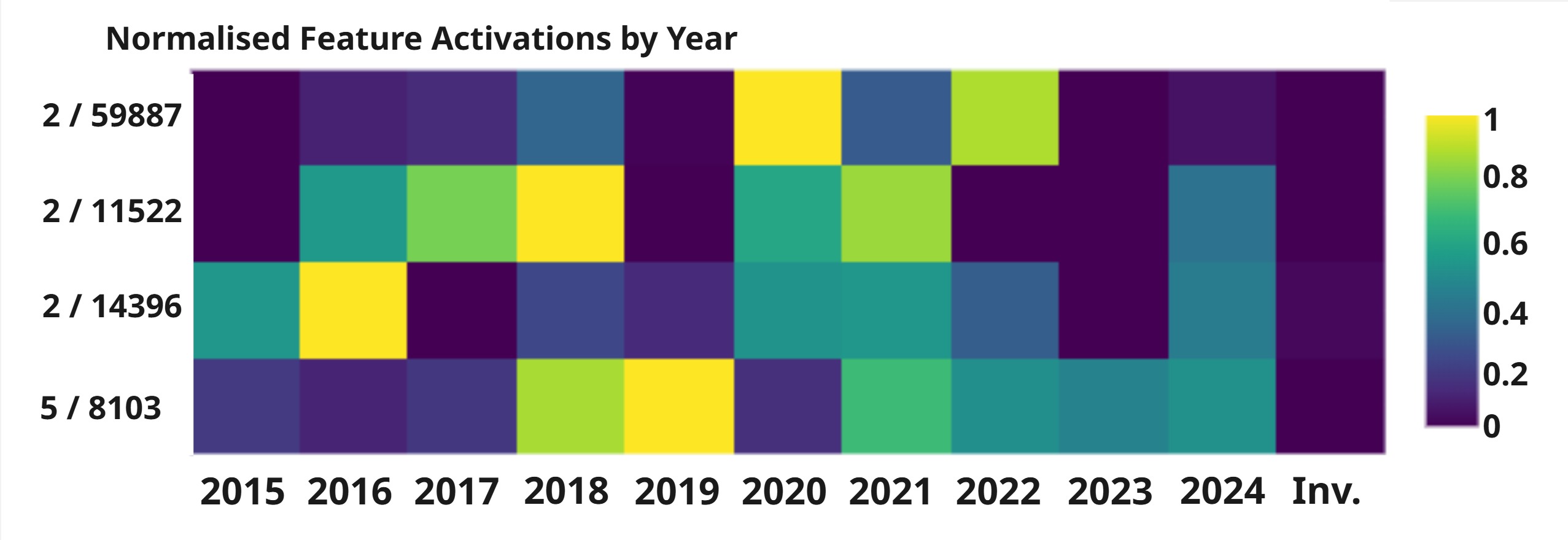}
  \caption{LLaMA 3.2 1B activation heatmap for chrono-semantic specialised features. The activation pattern is jagged, and not as smooth as Gemma 2 2B and Qwen3 4b across any span of time, potentially highlighting the lack of temporal consideration, or the limited ability of the transcoder to pick up on temporal features.}
  \label{fig:llama_activation_heatmap_chronosemantic}
\end{figure}

\section{Restoration studies}
Our initial findings for Gemma 2 2B use $\sim$31 nodes to ablate the understanding of time and revert the outcomes to a relevant and time invariant answer distribution. Whilst our temporal representation is quite small compared to the $\sim$17,000 nodes generated for every graph at the outset, we investigate whether a minimal subset of these feature nodes can achieve a similar outcome. Moreover, we want to understand if there is a more general representation of time that can help us ablate or steer for temporal outcomes.

As a starting point, we want to find the contribution of each \commontemporal{} node when removing the concept of explicit time. We zero ablate all \commontemporal{} except one and measure the explicit logit probability difference for relative and explicit answers. For Gemma 2 2B we find that a few nodes have a large contribution to explicit temporal recall, \texttt{L4F15166}, \texttt{L4F9816} and \texttt{L4F6098}. These nodes denote temporal concepts such as "numbers representing years",  "years starting with the number 20" and "years or judge, along with sometimes other numbers". Using the top 5 contributing nodes, we can dramatically shift the relative and explicit logit probabilities, whilst maintaining a substantially high F1 score. Testing against the next 5 nodes reveals contrasting results: explicit answers emerge (visible in F1 scores), with diminished \textit{new rve} improvement.

\begin{table}[h]
  \centering
  \resizebox{\columnwidth}{!}{%
  \begin{tabular}{lcccccccc}
    \textbf{Year} & \textbf{Ablation} & \textbf{Relative} & \textbf{Explicit} & \textbf{Explicit Logit Diff} & \textbf{Relative Logit Diff} & \textbf{Prior rve} & \textbf{New rve} \\
      \hline
        2020 & 0.562 & 0.514 & 0.908 & -0.078 & -0.023 & -0.052 & 0.003 \\
        2021 & 0.630 & 0.477 & 0.914 & -0.060 & -0.006 & -0.060 & -0.006 \\
        2022 & 0.620 & 0.425 & 0.917 & -0.066 & -0.011 & -0.052 & 0.003 \\
        2023 & 0.614 & 0.487 & 0.925 & -0.077 & -0.003 & -0.072 & 0.001 \\
        2024 & 0.672 & 0.549 & 0.903 & -0.015 & -0.004 & -0.011 & 0.000 \\
    \hline
  \end{tabular}%
  }
  \caption{Gemma 2 2B, Top 5 contributing \commontemporal{} nodes.}
  \label{tab:table_chqa_100_ablate_top5}
\end{table}

\section{Circuit tracing alternatives}
\subsection{EAP-IG Graphs}
\label{app:appendix_eapig_graphs}
\begin{figure*}[h]
  \centering 
  \includegraphics[width=1\linewidth]{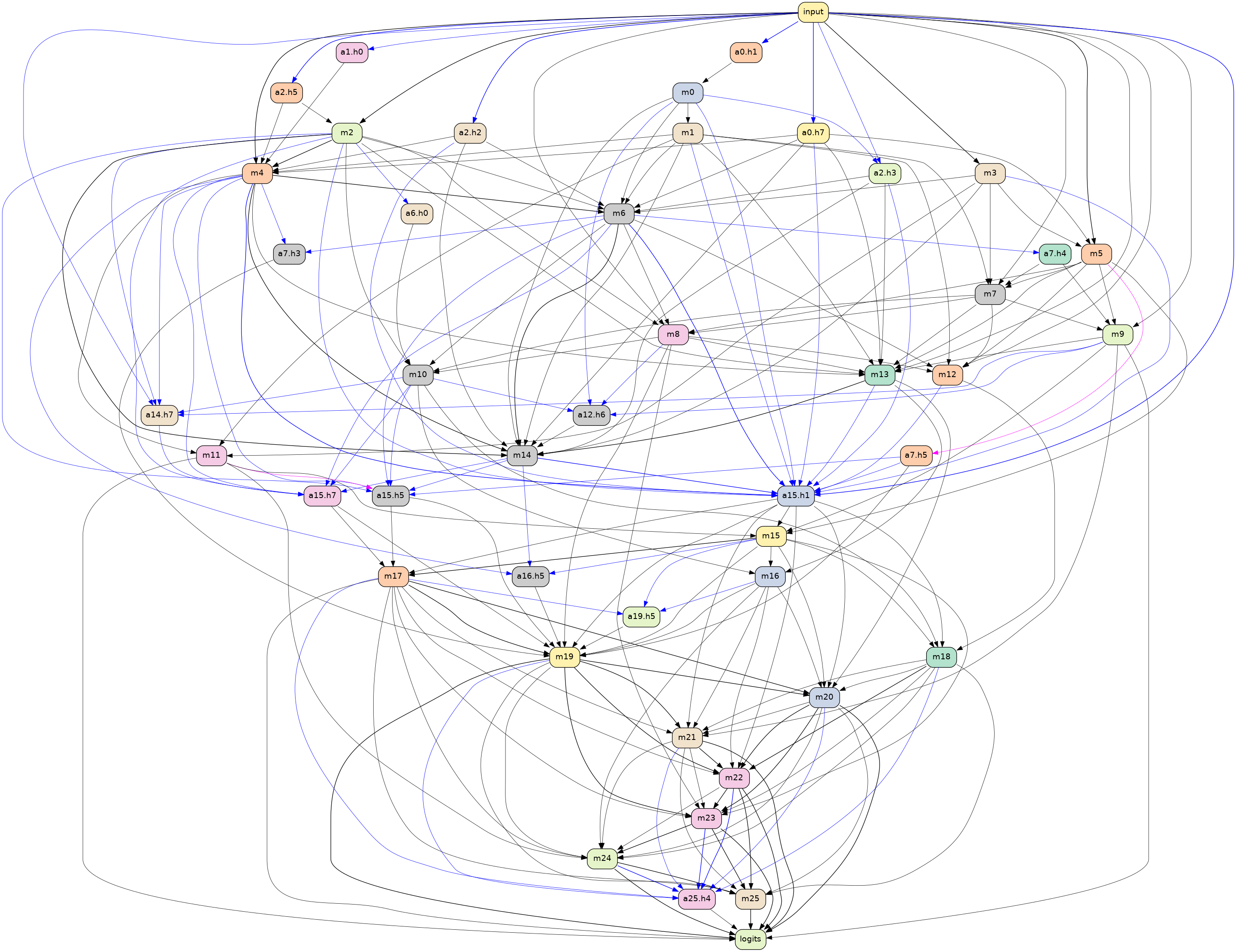}\hfill
  \caption{Gemma 2 sample EAP-IG circuit. Note the layer 7 head 4 attention, and M14 nodes}
  \label{fig:appendix_eapig_graphs}
\end{figure*}

As a complementary study to understand the overlap of understanding between LLM component circuit tracing and transcoder circuit tracing we recreate the temporal head study \citep{Park2025-ok} for Gemma 2 2B, Qwen3-4b and LLaMA 3.2 1B, upgrading the EAP-IG package to accommodate group query attention. Our fork of Temporal Heads with upgraded EAP-IG algorithm can be found here:  \url{https://github.com/sg-sy/TemporalHead}

\noindent\textbf{Gemma 2 2B:} Applying the methodology (Figure \ref{fig:appendix_eapig_graphs}) we find A0H7, A2H3 and A7H4 as a common attention temporal heads. Examining the MLP components we note that the M14 node is common across time invariant and time sensitive questions. However, as EAP-IG circuit analysis is based on neural components, we are unable to understand the features and the 'why?' behind the importance of M14.

\noindent\textbf{Qwen3-4b:} There are no apparent temporal attention heads for Qwen3.

\noindent\textbf{LLaMA 3.2 1B:}
There are no apparent temporal attention heads for LLaMA3.2.

\subsection{SAE Circuit Tracing}

\begin{figure*}[h]
  \centering 
  \includegraphics[width=1\linewidth]{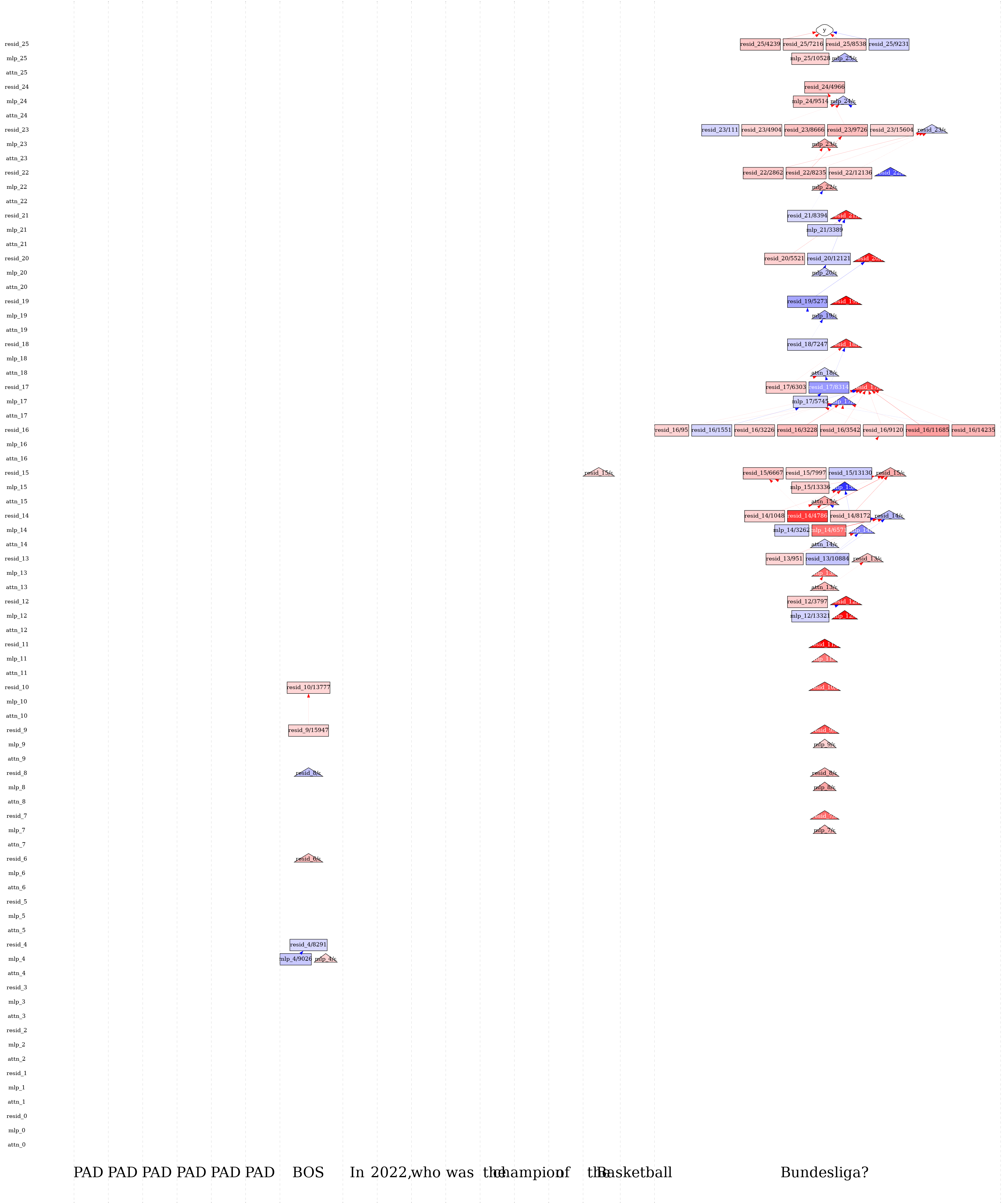}
  \caption{Gemma 2 2B SAE circuit tracing example showing residual stream and MLP feature interactions.}
  \label{fig:appendix_sae_graphs_gemma2}
\end{figure*}

We focussed on per layer transcoders as they produce richer representations and fewer error nodes leading to better interpretability and steering outcomes \citep{NEURIPS2024_2b8f4db0}. Examining a single question regarding the winner of the Basketball Bundesliga in 2022 and 2024 using Gemma 2 2B, SAEs (MLP, attention and residual stream saes) and Feature Circuits \citep{marks2025sparse}, we note a strong presence of residual stream and mlp features overlapping with our findings for higher temporal representations (Figure \ref{fig:appendix_sae_graphs_gemma2}). However, we also note a lack of nodes at lower layers, such as 4-7, which have been identified in our exploration of transcoders as containing both \commontemporal{} and \commontoyear{} nodes. Whilst we have conducted preliminary investigations into the overlap of Feature Circuit with Circuit tracing our exploration is limited. Our fork of Feature Circuits can be found here: \url{https://github.com/sg-sy/feature-circuits}

\section { Optimised circuit tracing}
\textbf{ Graphing differences for low confidence graphs}
Our initial conditions utilised high confidence outcomes, constructing circuits based on top-k logit outcomes. Whilst this provides a convenient way to interrogate for high confidence organic answers, for our purposes we want to understand the impact low confidence answers have on the influence scores, and the presentation of temporal nodes.

Low confidence graphs present similar year specific temporal features highlighting the strong contribution of temporal anchors even in low confidence settings. However, our selection of high confidence examples is justified if we examine the difference in temporal nodes between high and low confidence settings.

We note an increase in higher layer temporal representations (equal to or above layer 10) for our high confidence set, as well as an overall higher number of nodes. Surprisingly, average influence for all nodes per year across both sets are equivalent.

\begin{table}[h]
  \resizebox{\columnwidth}{!}{%
  \begin{tabular}{l|cc|cc}
     & \multicolumn{2}{c|}{\textbf{Overall Node Count}} & \multicolumn{2}{c}{\textbf{Higher Layer Node Count}} \\
    \hline
    \textbf{Year} & \textbf{High Conf.} & \textbf{Low Conf.} & \textbf{High Conf.} & \textbf{Low Conf.} \\
    \hline
    2020 & 56 & 37 & 6 & 1 \\
    2021 & 52 & 50 & 3 & 3 \\
    2022 & 44 & 46 & 5 & 3 \\
    2023 & 81 & 46 & 6 & 2 \\
    2024 & 106 & 64 & 4 & 1 \\
    Common & 28 & 30 & N/A & N/A\\
    \hline
  \end{tabular}%
  }
  \caption{Node counts by year comparing overall and high layer representations.}
  \label{tab:node_counts_by_year}
\end{table}

\subsection{Simplified Temporal Representations}

The indirect influence and node filtering from transcoder circuit tracing limit the number of nodes associated with the temporal component of our test statements. We want to further understand the overlap between our findings and simplified temporal representations. To do so, we generate a simplified version of an explicit prompt (such as "In 2020"). Our investigation finds that the simplified time statements for Gemma 2 2B produce a strong overlap with the high confidence graphs, but do not entirely eclipse their representations. Focussing on the layer 4-14, there are obvious differences. Discrepancies were observed between the node structures of the simplified and original prompts. Notably, certain expected nodes are absent. Most significant are nodes such as L6F2246; while categorised under 'years and dates,' it also activates on semantic markers like 'coronavirus' and 'Edo period,' suggesting a broader or more entangled encoding of temporal context.

\begin{figure}[h]
  \centering
  \includegraphics[width=0.8\linewidth]{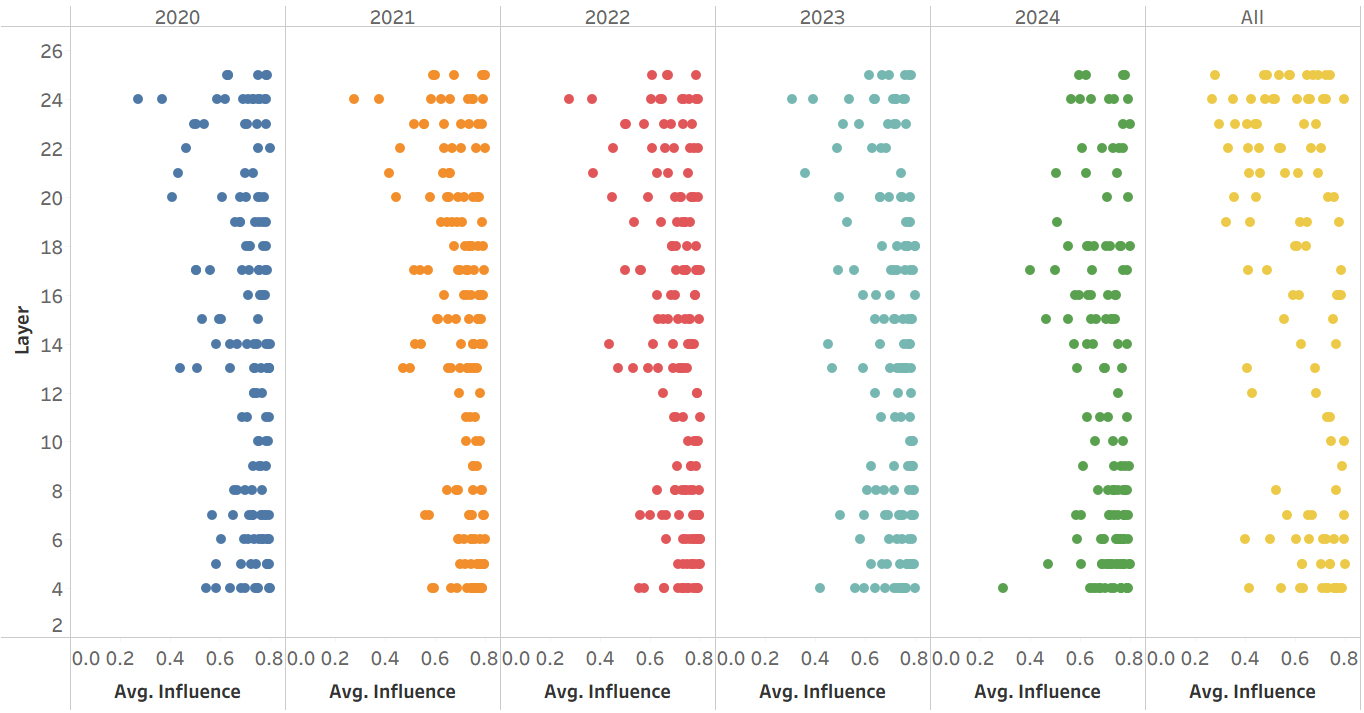}
  \caption{Simplified temporal representations for 2020--2024.}
  \label{fig:simplified_time_2020_2024}
\end{figure}

\section{Scaled Chrono-semantic Steering}
We conducted a preliminary investigation into scaled steering methodologies for improved steering outcomes. Given the partially overlapping and distinct temporal features of Gemma 2 2B and Qwen3-4b we hypothesised that a scaled representation of these features may lead to improved outcomes.

To explore scaled representations we first base lined a set of high confidence questions; the president of France and the United States of America over the period 2015 to 2024. We select these questions because of their high accuracy, and this period as it covers a change in presidents and overlaps with the training cutoff period of both models. Our baseline in this case is the original steering method. For comparison, we investigated two alternative methods.

\textbf{Average activation:} The application of average activation of Chrono-semantic nodes for the year we wish to align to, at layers 12 and above.

\textbf{Subtractive:} The addition of Chrono-semantic nodes for the year we wish to align, with a selection of Chrono-semantic nodes for the year we did \textbf{not} wish to align to. Both using layers 12 and above.

The results are mixed. In some respects the \textit{subtractive} intervention outperforms, producing marginally better results than the \textit{Average activation} intervention, and in others the opposite is true.

\begin{figure}[h]
  \centering
  \includegraphics[width=0.8\linewidth]{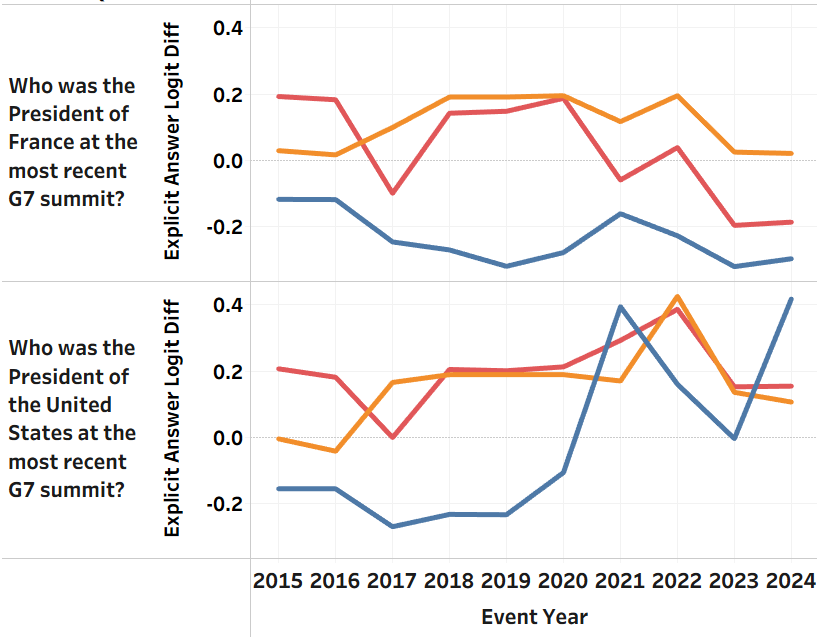}
  \caption{\textit{Original steering} (blue), \textit{average activation} (orange), \textit{subtractive} (red). Demonstrates the improvement of the explicit answer logit probabilities with a scaled or selective approach, when compared to the original steering methodology}
  \label{fig:scaled_steering_gemma}
\end{figure}

\noindent However, testing the \textit{subtractive} intervention on the wider ChronosAlign dataset produces suboptimal outcomes (Figure \ref{tab:table_chqa_100_scores_gemma2_scaled_steering_appendix} and \ref{tab:table_chqa_100_scores_gemma2_selective_steering_appendix}). This suggests there is a dynamic effect between these Chrono-semantic nodes and the question at hand. A future direction of research could consider the relationship between questions, activations and answer probabilities; the development of a learned scaled representation of these nodes for specific outcomes.

\begin{table}[ht!]
  \resizebox{\columnwidth}{!}{%
    \begin{tabular}{l|l|ccccc}
      \hline
       & \textbf{Metric} & \textbf{2020} & \textbf{2021} & \textbf{2022} & \textbf{2023} & \textbf{2024} \\
      \hline
      \multirow{4}{*}{F1} & Steer F1 & 0.189 & 0.197 & 0.142 & 0.194 & 0.202 \\
      & Relative F1 & 0.071 & 0.085 & 0.020 & 0.046 & 0.081 \\
      & Explicit F1 & 0.351 & 0.370 & 0.416 & 0.320 & 0.148 \\
      & Invariant F1 & 0.083 & 0.076 & 0.056 & 0.089 & 0.118 \\
      \hline
      \multirow{4}{*}{Logit Diff} & Explicit Logit Diff & 0.031 & 0.064 & 0.016 & 0.045 & 0.023 \\
      & Relative Logit Diff & -0.028 & -0.043 & -0.029 & -0.033 & -0.034 \\
      & Invariant Logit Diff & -0.033 & -0.050 & -0.031 & -0.038 & -0.037 \\
      & AE Logit Diff & 0.052 & 0.088 & 0.014 & 0.041 & 0.046 \\
      \hline
      \multirow{4}{*}{Probs. Change} & Prior ive & 0.073 & 0.079 & 0.076 & 0.074 & 0.045 \\
      & Prior rve & 0.045 & 0.052 & 0.048 & 0.047 & 0.018 \\
      & Steered ive & 0.009 & -0.035 & 0.028 & -0.009 & -0.015 \\
      & Steered rve & -0.014 & -0.055 & 0.003 & -0.032 & -0.039 \\
      \hline
      \end{tabular}%
  }
      \caption{F1 scores, with logit probability difference for invariant and relative answers. Gemma 2 2B, ChronosAlign dataset using \textit{Average activation} steering intervention with Chrono-semantic nodes at layer 12 and above.}
  \label{tab:table_chqa_100_scores_gemma2_scaled_steering_appendix}
\end{table}

\begin{table}[ht!]
  \resizebox{\columnwidth}{!}{%
    \begin{tabular}{l|l|ccccc}
      \hline
       & \textbf{Metric} & \textbf{2020} & \textbf{2021} & \textbf{2022} & \textbf{2023} & \textbf{2024} \\
      \hline
      \multirow{4}{*}{F1} & Steer F1 & 0.161 & 0.058 & 0.143 & 0.097 & 0.123 \\
      & Relative F1 & 0.071 & 0.085 & 0.020 & 0.046 & 0.081 \\
      & Explicit F1 & 0.351 & 0.370 & 0.416 & 0.320 & 0.148 \\
      & Invariant F1 & 0.083 & 0.076 & 0.056 & 0.089 & 0.118 \\
      \hline
      \multirow{4}{*}{Logit Diff} & Explicit Logit Diff & 0.036 & 0.015 & 0.042 & 0.012 & -0.000 \\
      & Relative Logit Diff & -0.013 & -0.019 & -0.027 & -0.018 & -0.018 \\
      & Invariant Logit Diff & -0.018 & -0.020 & -0.029 & -0.016 & -0.016 \\
      & AE Logit Diff & 0.029 & 0.015 & 0.035 & 0.006 & 0.005 \\
      \hline
      \multirow{4}{*}{Probs. Change} & Prior ive & 0.073 & 0.078 & 0.075 & 0.074 & 0.044 \\
      & Prior rve & 0.045 & 0.051 & 0.048 & 0.046 & 0.017 \\
      & Steered ive & 0.019 & 0.044 & 0.004 & 0.046 & 0.029 \\
      & Steered rve & -0.003 & 0.017 & -0.021 & 0.016 & -0.001 \\
      \hline
      \end{tabular}%
  }
      \caption{F1 scores, with logit probability difference for invariant and relative answers. Gemma 2 2B, ChronosAlign dataset using \textit{Subtractive} steering intervention with Chrono-semantic nodes, at layer 12 and above.}
  \label{tab:table_chqa_100_scores_gemma2_selective_steering_appendix}
\end{table}



\clearpage

\clearpage

\end{document}